\documentclass{article}

\usepackage{iclr2027_conference,times}
\iclrfinalcopy

\usepackage[T1]{fontenc}
\usepackage{amsmath,amssymb,bm}
\usepackage{graphicx}
\usepackage{booktabs,array}
\usepackage{microtype}
\usepackage[table]{xcolor}
\usepackage{xspace}
\usepackage{subcaption}
\usepackage{listings}
\usepackage{placeins}
\usepackage{hyperref}
\usepackage[capitalize,noabbrev]{cleveref}

\newcommand{\sG}{{\mathbb{G}}}
\newcommand{\sN}{{\mathbb{N}}}
\newcommand{\name}{{SchGen}}
\newcommand{\kicad}{\textit{KiCad}\xspace}
\newcommand{\ie}{i.e.}
\newcommand{\eg}{e.g.}

\lstdefinestyle{py}{
  language=Python,
  basicstyle=\ttfamily\footnotesize,
  keywordstyle=\bfseries\color{black!70!black},
  commentstyle=\itshape\color{gray!70},
  stringstyle=\color{green!40!black},
  showstringspaces=false,
  columns=fullflexible,
  frame=single,
  rulecolor=\color{gray!40},
  frameround=tttt,
  breaklines=true,
}

\newcommand{\codecell}[1]{%
  \begin{minipage}[c]{\linewidth}\ttfamily\tiny\raggedright
    \vspace*{2pt}%
    \setlength{\parindent}{0pt}#1
    \vspace*{2pt}%
  \end{minipage}%
}

\title{\name{}: PCB Schematic Generation with Semantic-Grounded Code Representations}

\author{%
  \begingroup\mdseries\begin{tabular}[t]{@{}p{0.47\textwidth}@{}p{0.47\textwidth}@{}}
  Qinpei Luo & Ruichun Ma\thanks{Ruichun Ma is the corresponding author.} \\
  University of California, San Diego & Microsoft Research Asia \\
  qpluo@ucsd.edu & ruichunma@microsoft.com \\[1.5ex]
  Xinyu Zhang & Lili Qiu \\
  University of California, San Diego &  The University of Texas at Austin \\
  xyzhang@ucsd.edu & lili@cs.utexas.edu \\
  \end{tabular}\endgroup%
}

\begin{document}

\maketitle

\begin{abstract}

Printed circuit board (PCB) schematic design defines nearly all electronic hardware, but it remains mostly manual and expertise-intensive.
While generative AI has advanced digital and analog IC design, PCB schematic generation from natural-language intent is largely unexplored.
This paper presents \name{}, a large language model that generates editable PCB schematics from natural-language requests.
The key challenge lies in the lack of an LLM-suited representation and a large-scale dataset.
Current schematic formats are dominated by verbose, tool-specific syntax and geometry-heavy descriptions, making them difficult to generate reliably.
We introduce a semantically grounded code representation that encodes schematic editing primitives with relative placement and pin-name-based wiring, transforming a geometry-driven generation problem into a semantics-driven matching task amenable to LLMs. We further construct a large-scale dataset of PCB schematics paired with user prompts via a human--agent collaborative pipeline that converts open-source hardware designs into our representation. Experiments show that \name{} significantly outperforms alternative representations and even larger general-purpose LLMs on wire connectivity accuracy and functional correctness.
Our results highlight the critical role of representation design in enabling generative models for complex hardware design tasks.
The source code is available at \url{https://github.com/microsoft/SchGen}.

\end{abstract}

\section{Introduction}
\label{intro}
Printed circuit boards (PCBs) are foundational to nearly all electronic hardware, from consumer electronics to Internet-of-Things devices and embodied AI hardware. As new applications proliferate, the demand for customized PCB designs is accelerating, while the design workflow remains largely manual, requiring substantial domain expertise to operate electronic design automation (EDA) tools.
Given a user's design request, engineers create the PCB board following the steps of: (1) \emph{Schematic Design}: Select circuit components (symbols) and specifying their interconnections (wires) which outputs a netlist file; (2) \emph{Layout}: Based on the netlist file, determine the placement of components on the metal layers and add metal wires/via holes; (3) \emph{Fabrication:} Produce the final Gerber file and send to factory to produce.
Among them, Schematic design is the critical and challenging step that defines the hardware architecture.
And we ask the question: \emph{how should schematic construction be represented to support reliable generation by language models?}

Recent works start to explore circuit schematic generation from natural-language requests using agent framework.
PCBSchemaGen~\citep{zou2026pcbschemagenconstraintguidedschematicdesign}
emphasizes constraint-guided iterative verification.
CircuitLM~\citep{hasan2026circuitlmmultiagentllmaideddesign}
combines component retrieval and staged reasoning to generate
custom CircuitJSON.
These training-free systems improve generation through domain knowledge prompting, structured intermediate formats, and inference-time orchestration.
However, they provide limited analysis of how representation affects the joint generation of component placement, spatial schematic layout, and net-level electrical connectivity.
In particular, the effects of such representation choices for model training remain unstudied.

\begin{figure*}[t]
  \setlength{\abovecaptionskip}{0pt}  %
  \centering
  \includegraphics[width=0.9\linewidth]{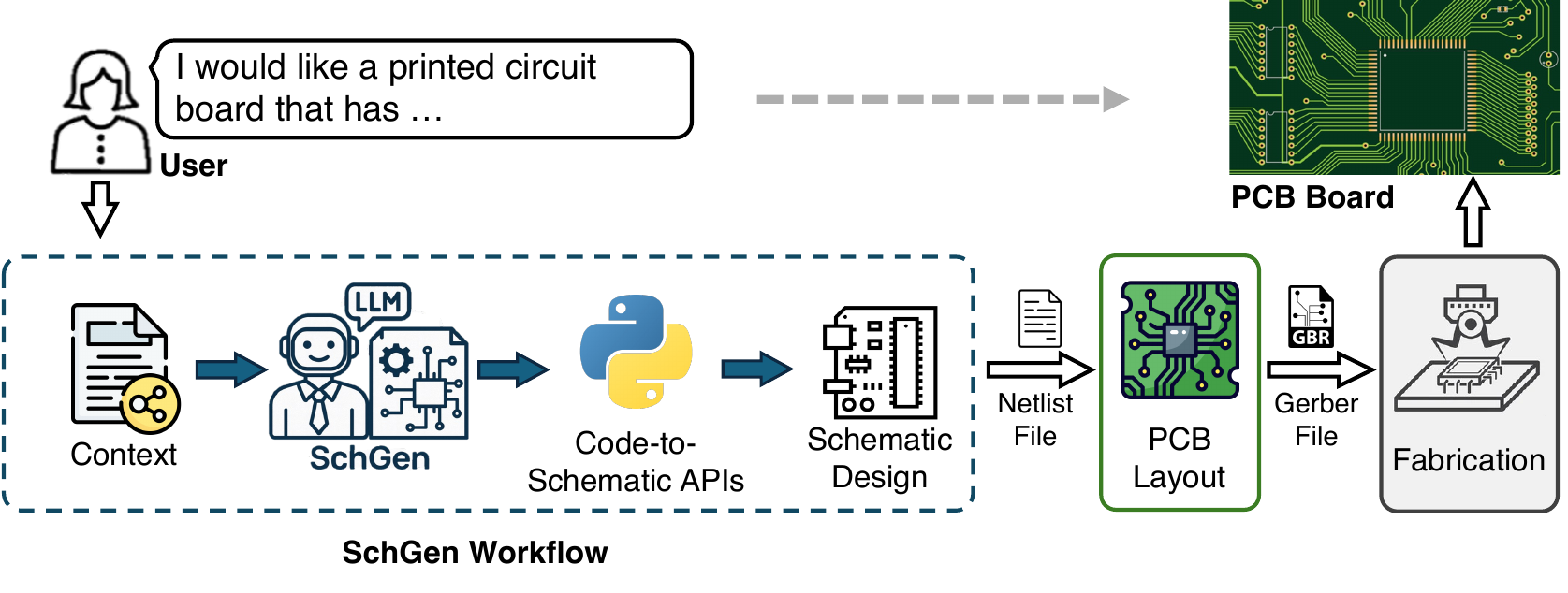}
  \caption{Overview of automated PCB design workflow. Based on the user request, \name{} generates a PCB schematic design using custom code representations, which is then converted to a netlist for PCB layout and fabrication.}
  \label{fig:overview}
\end{figure*}

\begin{figure}[t]
    \centering
    \includegraphics[width=0.9\linewidth]{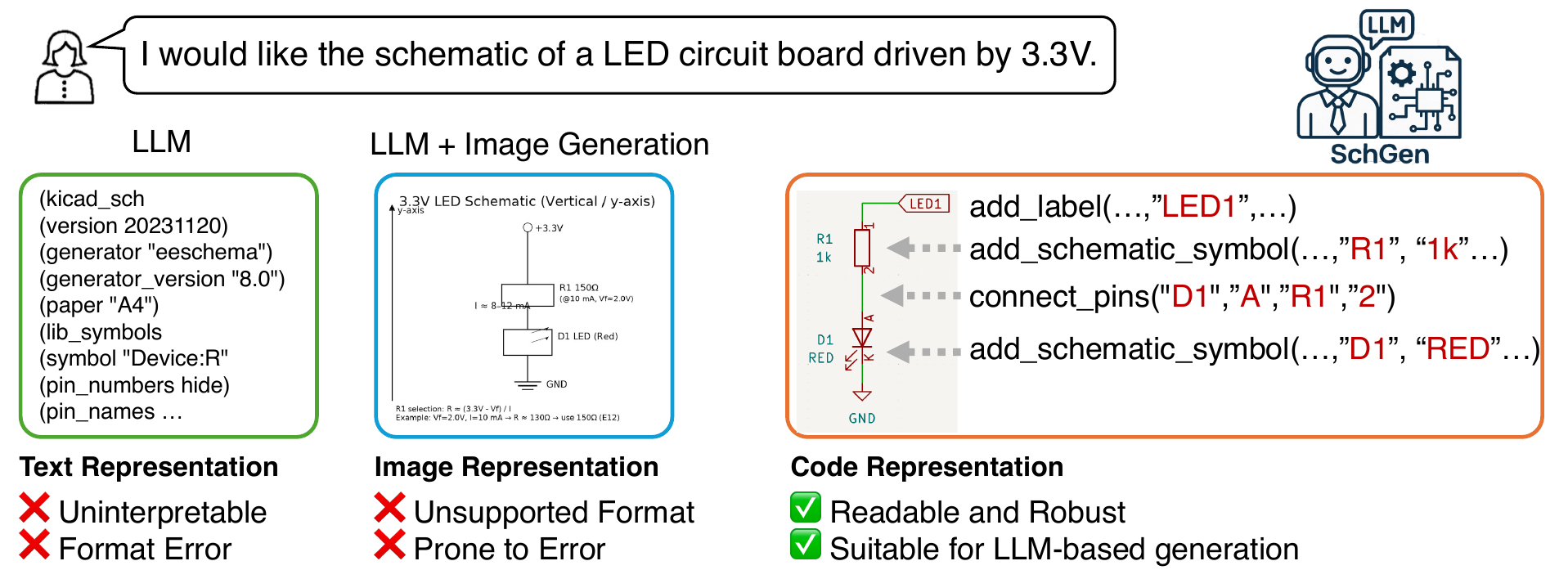}
    \caption{Comparison of schematic representations for LLM-based generation. Raw schematic files contain verbose metadata prone to formatting errors, while generated images are not machine-readable. Our code representation encodes editing primitives and pin-level semantics, transforming geometry prediction to a semantics-driven task amenable to LLM training.}
    \label{fig:comparison}
\end{figure}

We study the generation of editable PCB schematics from natural-language functional requirements.
Given a user's request, the goal is to produce a schematic
with correct electrical connectivity and a readable spatial
layout that engineers can inspect and modify in an EDA tool.
This problem presents two key challenges.
(i) \textbf{Representation}.
Unlike digital and analog IC design, where mature textual representations are
widely used, PCB schematic design remains largely centered around visual
operations such as selecting, placing, and connecting components.
As shown in \Cref{fig:comparison}, existing representations expose different
limitations for language-model generation.
Raw schematic files are dominated by verbose, tool-specific, and
version-dependent syntax and metadata, requiring the model to generate
low-level structural and geometric details in addition to circuit semantics.
Image-based representations preserve visual appearance but are neither
directly editable nor suitable for downstream electrical analysis.
Netlists encode electrical connectivity but do not themselves
specify component placement or wire geometry; these must be
supplied separately to obtain a readable schematic.
Consequently, generating an editable schematic requires a representation that
captures both electrical intent and spatial organization without forcing the
language model to explicitly predict unnecessary low-level geometry.
(ii) \textbf{Data scarcity}.
Although many open-source hardware designs are publicly available, they are distributed across various formats and repositories and are often not paired with natural-language design requirements or executable construction procedures, which makes them difficult to be reused for supervised training of
schematic-generation models.

To address the representation challenge, we introduce a semantically grounded executable representation of schematic construction which expresses component placement through offsets from local anchors and electrical connections through component references and pin names.
To address data scarcity, we develop an agentic pipeline that reconstructs open-source PCB designs as editable \kicad schematics, synthesizes corresponding natural-language requests, and converts each design into executable construction programs to form a dataset.

On top of the code representation, we build \name{}, a model for generating editable \kicad schematics from natural-language requests. Using the dataset, we fine-tune two open-weight backbones, GPT-OSS-20B~\citep{openai2025gptoss120bgptoss20bmodel} and Qwen3.8~\citep{qwen38}. We also provide the representation's API specification and examples to general LLMs for prompted generation, which allows us to examine representation choices under both supervised fine-tuning and prompting.

We evaluate \name{} across two fine-tuned backbones and prompted general-purpose LLMs, measuring execution success, netlist accuracy, spatial layout quality, and expert-assessed functional correctness. Representation ablations show that the proposed representation achieves the highest performance among the evaluated representations for both fine-tuned models with much smaller size compared to frontier models. Further generalization studies demonstrate that on an out-of-distribution dataset, fine-tuned \name{} still maintains its advanced ability and achieves the highest performance.

Our contributions are threefold:
(1) We introduce Code-L1, an executable schematic representation combining anchor-relative placement with named-pin connectivity, which we use for \name{} training.
(2) We construct a dataset of 2,105 schematics spanning 1,390 unique designs, paired with synthesized natural-language requests and executable construction programs through an agent-assisted pipeline with human correction.
(3) We evaluate representation choices across two fine-tuned backbones and prompted general-purpose LLMs, demonstrating improved netlist accuracy over alternative representations, meanwhile the finetuned small models achieve higher pass rate and netlist accuracy than general-purpose LLMs.

\section{Preliminaries and Related Work}
\label{related_works}

\subsection{PCB Schematic Design Process}
The PCB schematic design process conventionally depends on electronic design automation (EDA) software, including proprietary tools like \citet{Altium}, \citet{OrCAD}, \citet{robledo_future_2023}, and open-source alternatives like \citet{KiCad}.
While these tools support similar schematic content and editing operations, they rely on different file formats.
Throughout this paper, we use the \kicad schematic format because it is open source.
A schematic contains three main item types: (1) \emph{component symbols} representing physical circuit components such as chips, resistors, capacitors, (2) \emph{power symbols} representing rails such as \texttt{VCC} for power source, \texttt{GND} for ground, (3) \emph{net labels} that represent named net connections.
Labels with the same name are treated as electrically connected, and are commonly used to define module interfaces so that modules connect automatically via shared label names.
Each component symbol has one or more pins corresponding to the part's physical terminals; power symbols and net labels typically expose a single pin for wiring.
We reference pins by an ID or a concise name (e.g., \texttt{VCC}, \texttt{GND}, \texttt{GPIO1}), and draw wires between pins to specify electrical connections.

The schematic design workflow typically starts by selecting the required symbols, then placing them to form a readable layout, and finally wiring the pins to realize the intended circuit.
This process is critical as it decides hardware composition and electrical connections between every component, but it is also tedious as it involves placing dozens of symbols and connecting up to hundreds of pins.
However, current design automation mostly focuses on the next stage, PCB layout~\citep{9947957, 9529938, 10.1145/3626184.3635285} and routing~\citep{9264558, Li_Zhang_Xu_Liu_2023, siemens_design_2025}, while schematic design still heavily relies on labor-intensive, manual workflows.
SKiDL~\citep{skidl} uses programs to describe circuit components and electrical connections for netlist generation.

\subsection{Generative AI for Hardware Design}

Built on hierarchically abstracted Boolean logic representations, e.g., Verilog and VHDL, recent studies~\citep{wu2024chateda, thakur2024verigen, fu2023gpt4aigchip} have explored language model-based digital IC design.
For analog IC design,
CktGNN~\citep{dong2023cktgnn} first represents the circuit topology as graph structures, formulating a graph generation task to enable the design of various topologies.
AnalogCoder~\citep{lai2025analogcoder} presents a training-free LLM agent through Python code generation, while LaMAGIC~\citep{chang2024lamagic} fine-tunes a masked language model.
AnalogGenie~\citep{gao2025analoggenie} presents a comprehensive dataset to enable LLM pre-training for circuit topology generation.

Related work also studies translation between existing circuit representations. Netlistify~\citep{huang_netlistify} and Image2Net~\citep{xu2025image2netdatasetsbenchmarkhybrid} recover netlists from schematic images, while Schemato~\citep{matsuo2024schemato} converts netlists into LTSpice schematic files. These tasks support circuit digitization and visualization, but begin with an existing design whose connectivity is specified or recoverable from the input. Generating a schematic from a natural-language request additionally requires determining components and connections that satisfy the stated requirements.
\citet{zou2026pcbschemagenconstraintguidedschematicdesign} proposes a feedback-based iterative framework for PCB schematic design, while \citet{hasan2026circuitlmmultiagentllmaideddesign} introduces an agentic framework for LLM-aided circuit design. Our evaluation studies Code-L1 under both prompting and supervised fine-tuning.

\subsection{Hardware Design Datasets}
The development of AI-based hardware generation depends on training on large-volume and high-quality datasets.
\citet{8807057,dong2023cktgnn,gao2025analoggenie} provide datasets for analog circuits, but do not apply to PCB schematic design.
\citet{li2026pcbbench} provides a comprehensive benchmark designed to systematically evaluate LLMs, but it includes mostly PCB-related visual question answering which cannot be used for the design task.
Datasets for circuit topology generation and schematic conversion provide useful circuit structures or paired representations, but are not necessarily paired with natural-language requirements and executable schematic construction programs.
To supply the paired forms for supervised request-to-schematic generation, we have constructed a dataset of 1,390 unique schematic designs curated from open-source hardware designs on SparkFun Electronics~\citep{sparkfun_electronics} under the CC BY-SA 4.0 license.

\section{Approach}
\label{approach}

\name{} generates editable PCB schematics by translating natural-language requests into executable construction programs. We first propose the code representation named Code-L1 (Level 1), which specifies component placement and electrical connectivity through schematic editing operations.
We then present a pipeline for constructing paired requests, schematics, and programs, followed by the use of this representation for supervised fine-tuning and prompted generation.

\subsection{Code Representation for PCB Schematics}
\label{sec:representation}

Existing representations are hard for LLMs to learn from and generate correctly.
Although existing LLMs appear to have the ability to generate PCB schematic designs when asked in the prompt, they fail to produce valid schematic files that can be parsed by EDA tools, e.g., \kicad.
\Cref{fig:comparison} illustrates a comparison between existing representations and our proposed code representation.
When using text-based representation, \ie, raw \kicad schematic file's text content, LLMs struggle to capture the format of the schematic due to the presence of excessive schematic formatting details and redundant information that are irrelevant to the schematic's functionality, which also bring extra high token consumption.
When using LLM and image generation, the generated schematics contain distorted symbols and show a random format, making it infeasible to convert them into valid schematic files.

To tackle the challenges above, our goal is to find a new learning-efficient representation.
Our approach is inspired by the observation that human engineers typically follow a systematic process when drawing schematics that can be abstracted into a series of editing operations backed by clear rationales. We summarize two key insights from the above process:
(1) \emph{The schematic design can be abstracted as a series of editing operations, including adding symbols, placing labels, and connecting pins};
(2) \emph{The placement of symbols and labels is typically relative to a local reference based on functional correlations; the wire connections follow clear rationales based on the pin names that encode the pin functionalities.}
Specifically, engineers first place central symbols that represent the core circuit components, then arrange other symbols around them based on their functional connections.
Then, the pins are connected according to the connections between circuit component pins, \eg, \texttt{VCC} pins and \texttt{GND} pins are connected to power source and ground symbols, respectively.

Based on the observations, we introduce the following code APIs to form our code representation:
\begin{lstlisting}[style=py]
def add_schematic_symbol(symbol_lib, symbol_name, x, y, ref, value, rotation, mirror):
def add_label(label_pos, label_text, label_ref, label_type, text_orient):
def get_pin_location(symbol_ref, pin_name):
def connect_pins(symbol_a, pin_a, symbol_b, pin_b):
def write_out_all_wires():
\end{lstlisting}

\emph{add\_schematic\_symbol()} places a symbol with the given name from a symbol library on the assigned location with optional rotating and mirroring operations, meanwhile assigning a unique reference name and an optional value string.
\emph{add\_label()} places net labels with given text on specific locations and orientation, meanwhile allowing specifying label types (e.g., input, output, bidirectional) and a unique label reference ID.
\emph{get\_pin\_location()} gets the location of a specific pin, queried by symbol reference name and the pin name. For power symbols and net labels that have one pin only, the pin name is set as default to `1'.
\emph{connect\_pins()} connects two pins according to the symbol reference names and pin names of the two pins.
Finally, \emph{write\_out\_all\_wires()} writes out all wires to a \kicad schematic file with specified connections.

LLMs' ability to understand spatial relationships is relatively limited~\citep{yamada2024evaluatingspatialunderstandinglarge}.
For our task, the schematic may involve dozens of symbols/labels and numerical coordinates, which are hard for LLMs to generate correctly.
To handle it, we choose to use local coordinate instead of absolute coordinates in \emph{add\_schematic\_symbol} and \emph{add\_label}.
More specifically, we first get the coordinates of anchor points (for symbols, it is the center symbol in the circuit; for labels, it is the pin that it attaches to), then calculate the offsets and represent the coordinates with respect to the anchor points.
After executing the code, an editable valid \kicad schematic file is generated in \kicad tool, as illustrated in \Cref{fig:comparison}.
\Cref{app:portability} describes library and EDA-tool portability.

\begin{table*}[t]
\centering
\begingroup
\small
\setlength{\tabcolsep}{3pt}
\renewcommand{\arraystretch}{1.0}

\caption{Code representation illustration and ablation.}
\begin{tabular}{>{\centering\arraybackslash}m{0.13\textwidth}
                >{\centering\arraybackslash}m{0.56\textwidth}
                >{\centering\arraybackslash}m{0.08\textwidth}
                >{\centering\arraybackslash}m{0.08\textwidth}
                >{\centering\arraybackslash}m{0.08\textwidth}
                }
\toprule
Representation & Example & \text{MDL} & \text{LZ Norm} & \shortstack[c]{Val Loss\\($\times 10^{-2}$)} \\ \midrule
 &  & \multicolumn{3}{c}{(\text{mean} / \text{median})}\\
\midrule
\shortstack[c]{\textbf{Code-L1} \\ {\scriptsize(\name{})}} &
\codecell{add\_schematic\_symbol(...pos\_x=\textcolor{blue}{center\_x\_1}, pos\_y=\textcolor{blue}{center\_y\_1},...) \\
add\_schematic\_symbol(...pos\_x=\textcolor{blue}{center\_x\_1-58}, pos\_y=\textcolor{blue}{center\_y\_1+18},...) \\
\textcolor{blue}{connect\_pins("PWR1", "+1V8", "U2", "VDDIO")}
} & \textbf{2.19/2.31} & \textbf{1.36/1.42} & \textbf{1.29/0.25} \\
\midrule
\shortstack[c]{Code-L2 \\ {\scriptsize(w/o relative coord)}} &
\codecell{add\_schematic\_symbol(...pos\_x=\textcolor{blue}{157.48}, pos\_y=\textcolor{blue}{99.510},...) \\
add\_schematic\_symbol(...pos\_x=\textcolor{blue}{99.06}, pos\_y=\textcolor{blue}{117.29},...) \\
\textcolor{blue}{connect\_pins("PWR1", "+1V8", "U2", "VDDIO") }
} & 2.42/2.57 & 1.56/1.64 & 1.64/0.77 \\
\midrule
\shortstack[c]{Code-L3 \\ {\scriptsize(w/o pin name)}} &
\codecell{add\_schematic\_symbol(...pos\_x=\textcolor{blue}{157.48}, pos\_y=\textcolor{blue}{99.510},...) \\
add\_schematic\_symbol(...pos\_x=\textcolor{blue}{99.06}, pos\_y=\textcolor{blue}{117.29},...) \\
\textcolor{blue}{add\_new\_wire([99.06, 117.29], [114.3, 117.29])}
} & 2.43/2.63 & 1.60/1.64 & 3.96/3.20 \\
\bottomrule
\end{tabular}
\label{tab:representation_stat}
\endgroup
\end{table*}

\Cref{tab:representation_stat} shows the comparison between different schematic representations. \emph{Code-L2} replaces anchor-relative placement with absolute coordinates. \emph{Code-L3} additionally replaces named-pin connection operations with explicit wire segments. These variants examine the effects of relative placement and of delegating wire construction through electrical endpoint specification.

We use three metrics to compare these representations over our dataset:
(1) \textbf{Minimum Description Length (MDL)}: $\text{MDL} = {8\cdot \text{compressed\_bytes}}/{\text{raw\_bytes}}$, which uses lossless compression to approximate data complexity. Lower bits-per-byte indicates a more structured and learnable representation~\citep{NEURIPS2023_65a39213, shalev2014understanding}. %
(2) \textbf{Lempel--Ziv Complexity (LZ Norm)}: $\text{LZ\_norm}= {c(n)\,\log n}/{n}$, where $c(n)$ is the number of phrases from incremental parsing. A lower value implies lower intrinsic sequence complexity~\citep{maveli2026llmscompressanddecompress, 10.1162/089976604322860677}.
(3) \textbf{Validation Loss (Val Loss)}: We calculate the validation loss of each representation on the test dataset. Lower val loss indicates better performance and generalization ability. \Cref{app:loss-analysis} reports Pearson and Spearman correlations between validation loss and netlist accuracy at three model checkpoints.

Across the three code variants, Code-L1 achieves the lowest {MDL}, {LZ Norm}, and {Val Loss}, indicating that relative coordinates and pin-name connectivity jointly produce a more structured and compressible representation. These metrics serve as proxies for better learnability, a prediction we validate empirically in \Cref{sec:result}.  Our APIs abstract away the complexity of raw \kicad files by capturing key editing operations including relative-coordinate placement and pin-name-based connectivity, which enables structured and semantically meaningful schematic generation.

\begin{figure*}[t]
    \centering
    \includegraphics[width=0.9\linewidth]{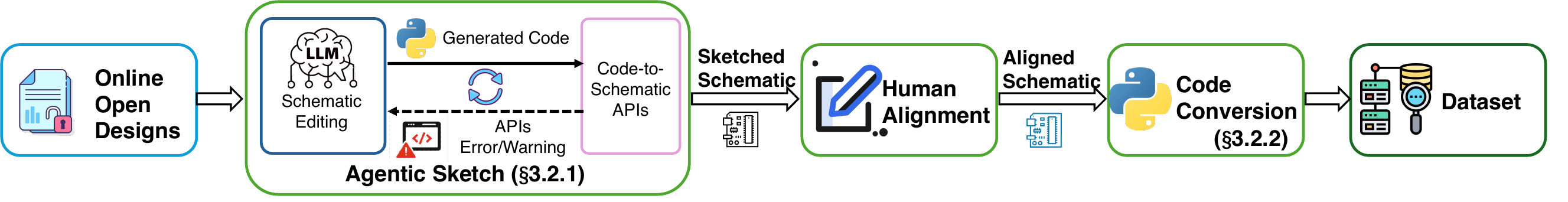}
    \caption{Pipeline of constructing the schematic dataset from online open-source PCB designs using LLM sketching and human correction.}
    \label{fig:dataset-pipeline}
\end{figure*}

\subsection{Constructing the Schematic Design Dataset}
\label{sec:data-pipeline}

Most online PCB designs provide an image of the schematic design instead of editable source files, which avoids issues with EDA tool compatibility and version discrepancies. However, it brings challenges to constructing a dataset based on schematic files.
Thus, we propose a pipeline shown in \Cref{fig:dataset-pipeline} to convert the design image to editable schematic and corresponding code representations.
We first introduce an \textit{agentic-sketch} module to acquire a draft~(\Cref{sketch}), which is later annotated by human engineers to fix possible errors.
Then, we develop a \textit{schematic-to-code} converter to translate the aligned schematic into code representations to form the dataset~(\Cref{code}).
Our approach enables scalable data acquisition from online open-source PCB designs.

\subsubsection{Agentic Sketch}
\label{sketch}

As shown in \Cref{fig:dataset-pipeline}, we take online images of reference designs as the input and leverage multi-modal LLMs, \eg, GPT-5, to generate Python code that uses APIs in \Cref{sec:representation}.
Through our underlying implementation, the compiled program outputs error and warning feedback in execution, e.g., wrong syntax and illegal symbols.
Then, the multi-modal LLM iterates code generation based on the feedback, until no error or reaching max iterations, and then outputs a sketch version of the schematic file by executing the code.

When sketching the schematics, multi-modal LLMs can make mistakes regarding complex wire connections even after iterations.
For example, it is hard for LLMs to determine whether two wires are connected or intersect.
Thus, we introduce manual editing to adjust the incorrect schematic designs to ensure their alignment with the online design reference. The agentic sketch can accurately reproduce symbols in \kicad from the image input within 5 rounds of iterations.
Combining the two steps, we produce a collection of PCB schematic designs in \kicad format by processing resources from online PCB designs like hardware datasheets and designs on \textit{SparkFun}.

\subsubsection{Schematic-to-Code Conversion}
\label{code}

Based on the collected schematic files, we develop a schematic-to-code converter that parses raw \kicad s-expression files into undirected graph representations, where pins and wires are treated as vertices and edges. Through graph traversal, we identify connected pin pairs and generate code following our API abstractions in \Cref{sec:representation}, including relative-coordinate symbol placement, label attachment, and pin-name-based connectivity, ultimately reproducing the original schematic with \emph{write\_out\_all\_wires}. A sample schematic and corresponding representations are shown in \Cref{example_code}.

Recall that our goal is to build a model that can one-shot generate a schematic based on the user request (\Cref{fig:overview}).
Thus, we synthesize user requests using external multi-modal LLMs, which take the exported images and netlists of schematics and generate requests from the user's perspective.
To further ensure consistency between the synthetic user requests and real ones, we prompt the model with examples authored by human users.
These reference examples include requests written by both PCB experts and non-experts; \Cref{app:request-style} reports results separately for the two request styles.
We also introduce two styles of requests to augment the dataset: concise and detailed, to model users with different background knowledge levels.
For the concise request, we assume the user has little knowledge of PCB design or only cares about high-level functionality; thus, the schematic is described with a brief summary of its function.
For the detailed request, the user request includes the specific circuit components and connections that the user wants to include in the schematic.
Two samples of different request styles of the schematic are shown in~\Cref{example_dataset}.

\subsection{Fine-Tuning and Prompted Generation}
\label{sec:training}

We separately fine-tune GPT-OSS-20B and Qwen3.8-27B to generate schematic construction programs from natural-language
requests. Both backbones use the same underlying request--program pairs, but receive backbone-specific synthetic chain-of-thought (CoT) supervision. Specifically, each backbone generates CoT traces for its own training examples, and these traces are incorporated into the corresponding training sequences. The CoT tokens contribute to the supervised training loss, providing explicit supervision for intermediate reasoning alongside schematic program generation.

At inference time, each model produces a construction program in a single generation attempt, which is executed to obtain an editable schematic. Detailed training settings are provided in \Cref{app:training}.

\section{Experimental Validation}
\label{result}

\subsection{Experiment Setup}
\label{sec:setup}

\textbf{Dataset details.}
Our proposed dataset contains 2,105 \kicad schematics from 1,390 unique designs. The total sample number is doubled to 4,210 using two styles of user request from two models as described in~\Cref{sec:training}. The schematics span various types of functionality, covering microcontroller, analog modules, LED, power, storage, battery, USB, antenna, connectors, etc., with up to 39 symbols and 48 labels per design. We partition the dataset at the schematic-design level to avoid possible leakage and randomly select 500 samples as the held-out test set, and use the remaining as
the training set.
For each model, we repeatedly run the experiments five times with different random seeds and calculate the corresponding standard deviation.

\textbf{Evaluation metrics.}
PCB schematics are typically evaluated from multiple perspectives, including rule-based methods~\citep{lee2003dispatching}, netlist verification~\citep{mitzner2009complete}, and design review by human experts. Based on these, we consider the following metrics:

\noindent\textbf{(a) Valid Circuits.}
Valid Circuits are determined by the ratio of generated circuits that can pass two sanity checks: (1) Python code can be successfully executed with no Python errors raised, which are typically triggered by incorrect arguments in the function calls, e.g., non-existent symbol reference or pin name, or coordinates out of range. (2) Zero critical errors reported by the Electrical Rules Check~(ERC) of \kicad, including short circuits, net conflicts, illegal connections, etc.

\noindent\textbf{(b) Spatial Violation.}
Spatial Violation measures the number of \textit{spatial overlaps} among symbols, labels, and wires. Each object is assigned a bounding box, and any intersection is counted as an overlap. To mitigate the statistical biases brought by different pass ratios, we normalize the average overlaps following: $\bar n_{weighted}=\frac{\bar n_{original}}{\text{pass ratio}}$. We use this metric as a readability proxy, as overlaps hinder engineer inspection and reflect the model's spatial reasoning ability.

\noindent\textbf{(c) Netlist Accuracy.}
Netlist Accuracy compares the symbols and connections of the generated netlist against the ground truth. We define a node $\boldsymbol{v}_i=(\text{symbol}, \text{pin})$; nodes sharing a net form a set $\sN$, and a netlist is $\sG=\{\sN_1,\cdots,\sN_m\}$. We report \emph{Jaccard}, \emph{Precision}, and \emph{Recall} between $\sG_{gen}$ and $\sG_{gt}$. Netlist accuracy is a strong proxy for schematic correctness, as it fully captures logical connectivity and serves as the sole input for downstream tasks of footprint layout and wiring.

\noindent\textbf{(d) Expert Verification.}
We have six experts evaluate the generated schematics using the rubric in \Cref{app:expert-rubric}. Symbol Error and Connection Error count incorrect/missing components and faulty pin connections (reported as averages). Functional Correctness reports the ratio of designs that can function as intended.\footnote{Cross-validation among six experts shows high agreement, with correctness ratio differences below 3\%.}
\Cref{app:checks-failures} reports the API checks, expert agreement, and failure cases.

\textbf{Baselines.}
To isolate the contribution of each design choice in our representation, we compare \name{}, trained with the proposed Code-L1 representation, against ablation variants: Code-L2 (w/o relative coordinates) and Code-L3 (w/o relative coordinates and pin-name connectivity), each finetuned from \emph{GPT-OSS-20B} and \emph{Qwen3.8-27B} on the same dataset.
To assess explicit reasoning supervision, we additionally train a \emph{CoT masked} variant for each backbone.
This variant retains the CoT traces in the training sequences but excludes their tokens from the supervised loss.
We compare the resulting models with prompted general-purpose LLMs.
We also include the raw \kicad schematic file representation, which directly uses the text of the \kicad schematic file as training data.
We compare \name{} with \emph{GPT-5.5}~\citep{gpt-5.5}, \emph{GPT-5.2}~\citep{gpt-5.2}, \emph{Grok-4}~\citep{Grok-4}, Claude Sonnet 5, Claude Haiku 4.5, and Gemini 3.8 Flash. 

\subsection{Main Results}
\label{sec:result}

\begin{table*}[t]
\centering
\caption{Performance comparison of \name{} finetuned with different base models and representations.}
\label{tab:code-eval}
\setlength{\tabcolsep}{2pt}
\renewcommand{\arraystretch}{1.0}
\footnotesize
\resizebox{\linewidth}{!}{%
\begin{tabular}{
l
>{\centering\arraybackslash}p{1.9cm}
>{\centering\arraybackslash}p{1.9cm}
>{\centering\arraybackslash}p{1.8cm}
>{\centering\arraybackslash}p{1.8cm}
>{\centering\arraybackslash}p{1.8cm}
>{\centering\arraybackslash}p{1.8cm}
>{\centering\arraybackslash}p{1.8cm}
>{\centering\arraybackslash}p{2.4cm}
}
\toprule
Method &
\multicolumn{1}{c}{Pass Rate $\uparrow$} &
\multicolumn{1}{c}{Spatial Violation $\downarrow$} &
\multicolumn{3}{c}{Netlist Accuracy (\%)$\uparrow$} &
\multicolumn{3}{c}{Expert Verification} \\
\cmidrule(lr){2-2} \cmidrule(lr){3-3} \cmidrule(lr){4-6} \cmidrule(lr){7-9}
& (Pass ratio\%) & (Number of overlaps)
& Jaccard & Precision & Recall
& Symbol Error $\downarrow$ & Connection Error $\downarrow$ & Functional Correctness(\%) $\uparrow$ \\
\midrule
\multicolumn{9}{l}{\textit{Qwen3.8-27B}} \\
\midrule
\textbf{Code-L1}
& $96.0 \pm 0.62$ & $\mathbf{5.08 \pm 0.12}$
& $\mathbf{70.7 \pm 0.58}$ & $\mathbf{76.3 \pm 0.65}$ & $\mathbf{74.5 \pm 0.61}$
& $\mathbf{0.02 \pm 0.01}$ & $\mathbf{0.24 \pm 0.03}$ & $\mathbf{82.6 \pm 0.54}$ \\

Code-L1 (CoT masked)
& $95.2 \pm 0.74$ & $5.43 \pm 0.15$
& $67.6 \pm 0.72$ & $73.2 \pm 0.80$ & $71.2 \pm 0.78$
& $0.05 \pm 0.01$ & $0.41 \pm 0.04$ & $76.0 \pm 0.61$ \\

Code-L2
& $87.6 \pm 0.95$ & $6.10 \pm 0.18$
& $50.9 \pm 0.88$ & $59.0 \pm 0.94$ & $54.0 \pm 0.85$
& $0.16 \pm 0.02$ & $1.15 \pm 0.08$ & $48.6 \pm 0.71$ \\

Code-L3
& $\mathbf{97.6 \pm 0.55}$ & $5.99 \pm 0.16$
& $19.6 \pm 0.42$ & $34.3 \pm 0.68$ & $20.3 \pm 0.45$
& $0.18 \pm 0.02$ & $4.92 \pm 0.22$ & $12.0 \pm 0.46$ \\

\kicad File
& $21.8 \pm 0.98$ & --
& $11.5 \pm 0.41$ & $13.4 \pm 0.45$ & $12.1 \pm 0.43$
& $1.15 \pm 0.07$ & $8.95 \pm 0.32$ & $4.60 \pm 0.30$ \\
\midrule
\multicolumn{9}{l}{\textit{GPT-OSS-20B}} \\
\midrule

Code-L1
& $82.0 \pm 0.86$ & $\mathbf{7.73 \pm 0.19}$
& $\mathbf{49.1 \pm 0.65}$ & $\mathbf{54.9 \pm 0.72}$ & $\mathbf{52.8 \pm 0.68}$
& $\mathbf{0.06 \pm 0.01}$ & $\mathbf{0.61 \pm 0.05}$ & $\mathbf{60.6 \pm 0.69}$ \\

Code-L1 (CoT masked)
& $53.4 \pm 1.12$ & $8.32 \pm 0.22$
& $30.5 \pm 0.84$ & $33.0 \pm 0.91$ & $35.4 \pm 0.88$
& $0.98 \pm 0.06$ & $1.58 \pm 0.09$ & $14.0 \pm 0.49$ \\

Code-L2
& $78.2 \pm 0.90$ & $7.77 \pm 0.17$
& $46.0 \pm 0.78$ & $52.6 \pm 0.85$ & $49.5 \pm 0.82$
& $0.28 \pm 0.03$ & $1.76 \pm 0.11$ & $33.0 \pm 0.67$  \\

Code-L3
& $\mathbf{86.4 \pm 0.94}$ & $8.14 \pm 0.24$
& $15.5 \pm 0.48$ & $24.8 \pm 0.62$ & $15.7 \pm 0.51$
& $0.24 \pm 0.03$ & $6.76 \pm 0.28$ & $6.00 \pm 0.34$ \\

\kicad File
& $32.5 \pm 1.25$ & $7.80 \pm 0.21$
& $9.11 \pm 0.35$  & $14.5 \pm 0.46$ & $9.33 \pm 0.38$
& $1.22 \pm 0.08$ & $9.42 \pm 0.35$ & $3.00 \pm 0.24$ \\

\bottomrule
\end{tabular}%
}
\end{table*}

\textbf{Effectiveness of code representations.}
\Cref{tab:code-eval} shows that Code-L1 achieves the highest netlist accuracy and expert-assessed functional correctness, with the lowest spatial violations, for both backbones. Functional correctness reaches 82.6\% for Qwen3.8-27B and 60.6\% for GPT-OSS-20B, compared with 48.6\% and 33.0\% for Code-L2. Replacing pin-based connections with explicit wire geometry further reduces functional correctness to 12.0\% and 6.00\% for Code-L3. Code-L1 checks named pins before constructing connections, whereas Code-L3 places wire segments directly from coordinates (\Cref{app:checks-failures}).

\textbf{Effect of CoT supervision.}
Masking the CoT loss reduces netlist Jaccard from 70.7\% to 67.6\% for Qwen3.8-27B and from 49.1\% to 30.5\% for GPT-OSS-20B. Functional correctness decreases from 82.6\% to 76.0\% and from 60.6\% to 14.0\%, respectively. Explicit CoT supervision therefore benefits both backbones, with a larger improvement for GPT-OSS-20B. Since reasoning traces remain in the training sequences, this ablation measures the contribution of directly supervising CoT in the reasoning process over schematic connectivity logic.

\begin{table*}[t]
\centering
\setlength{\belowcaptionskip}{0pt}
\caption{Performance comparison of different models. Bold marks the best values within each panel.}
\label{tab:benchmark}
\label{tab:additional-models}
\setlength{\tabcolsep}{2pt}
\renewcommand{\arraystretch}{1.15}
\footnotesize
\resizebox{\linewidth}{!}{%
\begin{tabular}{
l
>{\centering\arraybackslash}p{1.9cm}
>{\centering\arraybackslash}p{1.9cm}
>{\centering\arraybackslash}p{1.8cm}
>{\centering\arraybackslash}p{1.8cm}
>{\centering\arraybackslash}p{1.8cm}
>{\centering\arraybackslash}p{1.8cm}
>{\centering\arraybackslash}p{1.8cm}
>{\centering\arraybackslash}p{2.4cm}
}
\toprule
Method &
\multicolumn{1}{c}{Pass Rate $\uparrow$} &
\multicolumn{1}{c}{Spatial Violation $\downarrow$} &
\multicolumn{3}{c}{Netlist Accuracy (\%)$\uparrow$} &
\multicolumn{3}{c}{Expert Verification} \\
\cmidrule(lr){2-2} \cmidrule(lr){3-3} \cmidrule(lr){4-6} \cmidrule(lr){7-9}
& (Pass ratio\%) & (Number of overlaps)
& Jaccard & Precision & Recall
& Symbol Error $\downarrow$ & Connection Error $\downarrow$ & Functional Correctness(\%) $\uparrow$ \\
\midrule
\textbf{SchGen(Qwen3.8-27B-L1)}
& $\mathbf{96.0 \pm 0.62}$ & $\mathbf{5.08 \pm 0.12}$
& $\mathbf{70.7 \pm 0.58}$ & $\mathbf{76.3 \pm 0.65}$ & $\mathbf{74.5 \pm 0.61}$
& $\mathbf{0.02 \pm 0.01}$ & $\mathbf{0.24 \pm 0.03}$ & $\mathbf{82.6 \pm 0.54}$ \\

\textbf{SchGen(GPT-OSS-20B-L1)}
& $\mathbf{82.0 \pm 0.86}$ & $\mathbf{7.73 \pm 0.19}$
& $\mathbf{49.1 \pm 0.65}$ & $\mathbf{54.9 \pm 0.72}$ & $\mathbf{52.8 \pm 0.68}$
& $\mathbf{0.06 \pm 0.01}$ & $\mathbf{0.61 \pm 0.05}$ & $\mathbf{60.6 \pm 0.69}$ \\

\hline
GPT-5.5-L1
& $\mathbf{64.8 \pm 0.92}$ & $8.13 \pm 0.18$
& $\mathbf{45.8 \pm 0.74}$ & $\mathbf{49.1 \pm 0.80}$ & $\mathbf{48.6 \pm 0.77}$
& $\mathbf{0.08 \pm 0.01}$ & $\mathbf{0.82 \pm 0.06}$ & $\mathbf{55.0 \pm 0.70}$ \\

GPT-5.5-L2
& $62.0 \pm 0.98$ & $16.4 \pm 0.34$
& $40.9 \pm 0.82$ & $44.2 \pm 0.86$ & $43.8 \pm 0.85$
& $0.30 \pm 0.03$ & $2.10 \pm 0.12$ & $31.0 \pm 0.65$ \\

GPT-5.5-L3
& $58.6 \pm 1.05$ & $\mathbf{7.30 \pm 0.17}$
& $37.2 \pm 0.76$ & $40.4 \pm 0.81$ & $40.6 \pm 0.79$
& $0.26 \pm 0.03$ & $6.20 \pm 0.26$ & $10.6 \pm 0.43$ \\

GPT-5.5-\kicad
& $12.8 \pm 0.85$ & --
& $5.87 \pm 0.28$ & $6.57 \pm 0.31$ & $6.24 \pm 0.29$
& $1.35 \pm 0.08$ & $9.80 \pm 0.36$ & $2.00 \pm 0.20$ \\

\hline
GPT-5.2-L1
& $67.9 \pm 0.88$ & $\mathbf{8.56 \pm 0.19}$
& $\mathbf{43.0 \pm 0.71}$ & $\mathbf{45.7 \pm 0.76}$ & $\mathbf{49.5 \pm 0.78}$
& $\mathbf{0.11 \pm 0.02}$ & $\mathbf{0.94 \pm 0.07}$ & $\mathbf{50.0 \pm 0.71}$ \\

GPT-5.2-L2
& $59.3 \pm 1.02$ & $11.1 \pm 0.26$
& $35.5 \pm 0.79$ & $37.5 \pm 0.83$ & $40.1 \pm 0.85$
& $0.34 \pm 0.03$ & $2.45 \pm 0.14$ & $27.6 \pm 0.63$ \\

GPT-5.2-L3
& $\mathbf{68.0 \pm 0.86}$ & $9.55 \pm 0.22$
& $41.1 \pm 0.75$ & $43.4 \pm 0.80$ & $46.7 \pm 0.82$
& $0.28 \pm 0.03$ & $6.85 \pm 0.28$ & $9.00 \pm 0.41$ \\

GPT-5.2-\kicad
& $10.5 \pm 0.76$ & --
& $0.53 \pm 0.12$ & $0.88 \pm 0.15$ & $0.88 \pm 0.15$
& $1.48 \pm 0.09$ & $10.5 \pm 0.38$ & $1.00 \pm 0.14$ \\

\hline
Grok-4-L1
& $\mathbf{61.2 \pm 0.95}$ & $\mathbf{8.82 \pm 0.21}$
& $\mathbf{44.0 \pm 0.73}$ & $\mathbf{49.8 \pm 0.82}$ & $\mathbf{48.7 \pm 0.79}$
& $\mathbf{0.20 \pm 0.02}$ & $\mathbf{1.19 \pm 0.08}$ & $\mathbf{47.0 \pm 0.71}$ \\

Grok-4-L2
& $52.1 \pm 1.08$ & $10.7 \pm 0.25$
& $32.7 \pm 0.76$ & $34.6 \pm 0.80$ & $34.1 \pm 0.78$
& $0.42 \pm 0.04$ & $2.85 \pm 0.16$ & $23.6 \pm 0.60$ \\

Grok-4-L3
& $53.1 \pm 1.04$ & $9.98 \pm 0.24$
& $35.9 \pm 0.78$ & $37.8 \pm 0.83$ & $37.3 \pm 0.81$
& $0.36 \pm 0.03$ & $7.42 \pm 0.31$ & $7.60 \pm 0.37$ \\

Grok-4-\kicad
& $3.95 \pm 0.45$ & --
& $1.32 \pm 0.18$ & $1.32 \pm 0.18$ & $1.32 \pm 0.18$
& $1.65 \pm 0.10$ & $11.2 \pm 0.41$ & $0.60 \pm 0.11$ \\

\hline
Claude-Sonnet-5-L1
& $\mathbf{93.4 \pm 0.68}$ & $\mathbf{5.42 \pm 0.14}$
& $68.0 \pm 0.60$ & $73.7 \pm 0.66$ & $71.2 \pm 0.63$
& $\mathbf{0.03 \pm 0.01}$ & $\mathbf{0.31 \pm 0.03}$ & $\mathbf{78.4 \pm 0.58}$ \\

Claude-Sonnet-5-L2
& $91.8 \pm 0.72$ & $7.85 \pm 0.19$
& $\mathbf{68.3 \pm 0.59}$ & $\mathbf{73.7 \pm 0.65}$ & $\mathbf{71.4 \pm 0.62}$
& $0.14 \pm 0.02$ & $1.15 \pm 0.08$ & $54.2 \pm 0.69$ \\

Claude-Sonnet-5-L3
& $76.5 \pm 0.89$ & $6.90 \pm 0.16$
& $52.0 \pm 0.68$ & $57.0 \pm 0.74$ & $54.6 \pm 0.71$
& $0.19 \pm 0.02$ & $4.85 \pm 0.22$ & $18.4 \pm 0.54$ \\

Claude-Sonnet-5-\kicad
& $48.2 \pm 0.96$ & --
& $35.2 \pm 0.75$ & $43.9 \pm 0.81$ & $37.3 \pm 0.77$
& $0.68 \pm 0.05$ & $6.90 \pm 0.29$ & $14.2 \pm 0.49$ \\

\hline
Claude-Haiku-4.5-L1
& $\mathbf{91.2 \pm 0.74}$ & $\mathbf{5.85 \pm 0.15}$
& $\mathbf{67.0 \pm 0.61}$ & $\mathbf{72.1 \pm 0.67}$ & $\mathbf{70.4 \pm 0.64}$
& $\mathbf{0.04 \pm 0.01}$ & $\mathbf{0.36 \pm 0.04}$ & $\mathbf{75.8 \pm 0.60}$ \\

Claude-Haiku-4.5-L2
& $84.6 \pm 0.83$ & $9.12 \pm 0.21$
& $62.7 \pm 0.64$ & $69.1 \pm 0.70$ & $66.5 \pm 0.67$
& $0.18 \pm 0.02$ & $1.42 \pm 0.09$ & $48.6 \pm 0.70$ \\

Claude-Haiku-4.5-L3
& $44.5 \pm 1.06$ & $10.4 \pm 0.25$
& $26.7 \pm 0.78$ & $31.4 \pm 0.83$ & $27.3 \pm 0.79$
& $0.45 \pm 0.04$ & $7.80 \pm 0.32$ & $6.40 \pm 0.34$ \\

Claude-Haiku-4.5-\kicad
& $18.4 \pm 0.88$ & --
& $9.06 \pm 0.34$ & $13.4 \pm 0.41$ & $11.7 \pm 0.38$
& $1.18 \pm 0.07$ & $9.10 \pm 0.34$ & $3.20 \pm 0.25$ \\

\hline
Gemini-3.8-Flash-L1
& $88.5 \pm 0.78$ & $\mathbf{6.10 \pm 0.15}$
& $63.2 \pm 0.63$ & $66.5 \pm 0.69$ & $65.8 \pm 0.66$
& $\mathbf{0.04 \pm 0.01}$ & $\mathbf{0.42 \pm 0.04}$ & $\mathbf{72.4 \pm 0.62}$ \\

Gemini-3.8-Flash-L2
& $\mathbf{90.4 \pm 0.75}$ & $8.34 \pm 0.20$
& $\mathbf{66.4 \pm 0.61}$ & $\mathbf{70.1 \pm 0.67}$ & $\mathbf{69.3 \pm 0.65}$
& $0.15 \pm 0.02$ & $1.28 \pm 0.09$ & $51.8 \pm 0.70$ \\

Gemini-3.8-Flash-L3
& $86.2 \pm 0.81$ & $7.45 \pm 0.18$
& $63.5 \pm 0.64$ & $66.7 \pm 0.69$ & $66.8 \pm 0.67$
& $0.16 \pm 0.02$ & $3.95 \pm 0.19$ & $24.6 \pm 0.60$ \\

Gemini-3.8-Flash-\kicad
& $74.8 \pm 0.90$ & --
& $58.7 \pm 0.66$ & $61.5 \pm 0.71$ & $61.7 \pm 0.69$
& $0.32 \pm 0.03$ & $4.40 \pm 0.21$ & $28.4 \pm 0.63$ \\

\bottomrule
\end{tabular}%
}
\end{table*}

\textbf{Comparison with general-purpose models.}
\Cref{tab:benchmark} shows that \name{} with Qwen3.8-27B achieves the best reported results across all metrics on the in-domain benchmark, including a 96.0\% pass rate, 70.7\% netlist Jaccard, and 82.6\% functional correctness. Compared with GPT-5.5 using Code-L1, it improves Jaccard and functional correctness by 24.9 and 27.6 percent, respectively.
\Cref{app:additional-models} summarizes the Claude and Gemini comparisons.

\textbf{Effectiveness under prompting.}
Code-L1 benefits prompted generation as well as supervised fine-tuning. With API specifications and examples supplied in the prompt, it improves expert-assessed functional correctness over Code-L2 by 22.4--24.0 percent across GPT-5.5, GPT-5.2, and Grok-4.

\begin{table}[t]
\centering
\caption{Performance on the OOD GitHub dataset. Bold marks the best values within each panel.}
\label{tab:quantitative_generalization}
\setlength{\tabcolsep}{2pt}
\renewcommand{\arraystretch}{1.15}
\footnotesize
\resizebox{\linewidth}{!}{%
\begin{tabular}{
l
>{\centering\arraybackslash}p{1.9cm}
>{\centering\arraybackslash}p{1.9cm}
>{\centering\arraybackslash}p{1.8cm}
>{\centering\arraybackslash}p{1.8cm}
>{\centering\arraybackslash}p{1.8cm}
>{\centering\arraybackslash}p{1.8cm}
>{\centering\arraybackslash}p{1.8cm}
>{\centering\arraybackslash}p{2.4cm}
}
\toprule
Method &
\multicolumn{1}{c}{Pass Rate $\uparrow$} &
\multicolumn{1}{c}{Spatial Violation $\downarrow$} &
\multicolumn{3}{c}{Netlist Accuracy (\%)$\uparrow$} &
\multicolumn{3}{c}{Expert Verification} \\
\cmidrule(lr){2-2} \cmidrule(lr){3-3} \cmidrule(lr){4-6} \cmidrule(lr){7-9}
& (Pass ratio\%) & (Number of overlaps)
& Jaccard & Precision & Recall
& Symbol Error $\downarrow$ & Connection Error $\downarrow$ & Functional Correctness(\%) $\uparrow$ \\
\midrule
\textbf{\name{}}(Qwen3.8-27B)
& $\mathbf{90.7 \pm 0.72}$ & $4.55 \pm 0.11$
& $\mathbf{56.7 \pm 0.65}$ & $\mathbf{63.5 \pm 0.70}$ & $\mathbf{60.3 \pm 0.68}$
& $\mathbf{0.04 \pm 0.01}$ & $\mathbf{0.38 \pm 0.04}$ & $\mathbf{71.6 \pm 0.64}$ \\

\textbf{\name{}}(GPT-OSS-20B)
& $83.2 \pm 0.84$ & $\mathbf{3.63 \pm 0.09}$
& $44.2 \pm 0.72$ & $50.9 \pm 0.78$ & $47.7 \pm 0.74$
& $0.07 \pm 0.01$ & $0.68 \pm 0.05$ & $56.0 \pm 0.70$ \\

GPT-5.2
& $62.1 \pm 0.90$ & $8.72 \pm 0.20$
& $40.5 \pm 0.74$ & $44.2 \pm 0.79$ & $42.7 \pm 0.76$
& $0.14 \pm 0.02$ & $1.06 \pm 0.08$ & $46.6 \pm 0.71$ \\
\bottomrule
\end{tabular}%
}
\end{table}

\textbf{Generalization.}
We further test whether \name{} can generate novel schematics that are out of the domain of our training set.
For this purpose, we build a challenging out-of-distribution~(OOD) dataset, which consists of 988 samples collected from 20 open-source \kicad projects on GitHub, which are from a completely different source compared to the dataset used for training. As shown in \Cref{tab:quantitative_generalization}, both \name{} implementations outperform GPT-5.2 across all reported metrics. The Qwen3.8-27B implementation achieves a 90.7\% pass rate, 56.7\% netlist Jaccard, and 71.6\% expert-assessed functional correctness, compared with GPT-5.2's 62.1\%, 40.5\%, and 46.6\%, respectively. It also produces fewer component and connection errors, indicating that its advantage extends beyond execution success to functional quality. The GPT-OSS-20B implementation achieves 56.0\% functional correctness and the fewest spatial violations among the evaluated models. These results support the transfer of \name{} to schematics collected outside the training-data source, with improvements in connectivity, readability, and expert-assessed functionality on the evaluated GitHub designs.
Two hardware-design case studies and design-time comparisons are presented in \Cref{app:hardware-cases}.

\section{Conclusion}

In this work, we introduced \name{}, a large language model for PCB schematic design generation.
By proposing a semantic-grounded code representation, we transform schematic design into a structured sequence of interpretable editing Python code, which allows LLMs to effectively capture both spatial reasoning and functional intent.
We further constructed a dataset of schematic designs through an agent--human collaborative pipeline, providing high-quality training data for the new task.
Experimental results demonstrate that \name{} substantially outperforms ablation variants and general-purpose models in terms of valid circuit generation, spatial arrangement, and netlist accuracy.
These findings demonstrate representation design as the key enabler for hardware design generation.

\clearpage
\subsection*{AI Use Statement}
We used LLMs to assist with language and LaTeX polishing, and synthetic dataset construction. The authors take full responsibility for the final manuscript, methods, results, and claims.

\subsection*{Ethics Statement}

\name{} aims to make PCB schematic design more accessible, while recognizing that errors in generated circuits can cause hardware failures or physical harm. The reported execution checks, connectivity metrics, and expert assessments do not establish safety or reliability in physical deployment. Generated designs therefore require qualified engineering review and application-appropriate electrical validation before fabrication or use, particularly in safety-critical settings. The dataset is derived from publicly available hardware designs, including SparkFun designs described in the paper as licensed under CC BY-SA 4.0; redistribution of source and derived assets must preserve applicable attribution and license obligations. The study does not involve external human participants or the collection of personal or sensitive data; expert evaluation was performed by co-authors and external experts. The concentration on SparkFun designs and the use of synthetic requests may limit coverage of circuit families and real user requirements, so the reported results should be interpreted within the evaluated scope.

\subsection*{Reproducibility Statement}

We document the schematic representation and editing APIs in \Cref{sec:representation}, the reconstruction and conversion pipeline in \Cref{sec:data-pipeline}, and request synthesis, reasoning-trace generation, and model training in \Cref{sec:training}. \Cref{sec:setup} describes the dataset split, evaluation metrics, expert assessment procedure, and baseline setup. The appendix provides schematic and code examples (\Cref{example_code}), examples of both request styles (\Cref{example_dataset}), dataset statistics (\Cref{dataset:stat}), training hyperparameters, and compute details (\Cref{app:training}).
Supplementary evaluations explicitly distinguish single-run results, benchmark reference statistics, and descriptive workflow observations.

\bibliographystyle{iclr2027_conference}
\bibliography{bibs}

\newpage
\appendix
\section{Appendix}

\subsection{Example of Schematic Design and Code Representations}
\label{example_code}

\begin{figure}[htbp]
    \centering
    \includegraphics[width=0.65\linewidth]{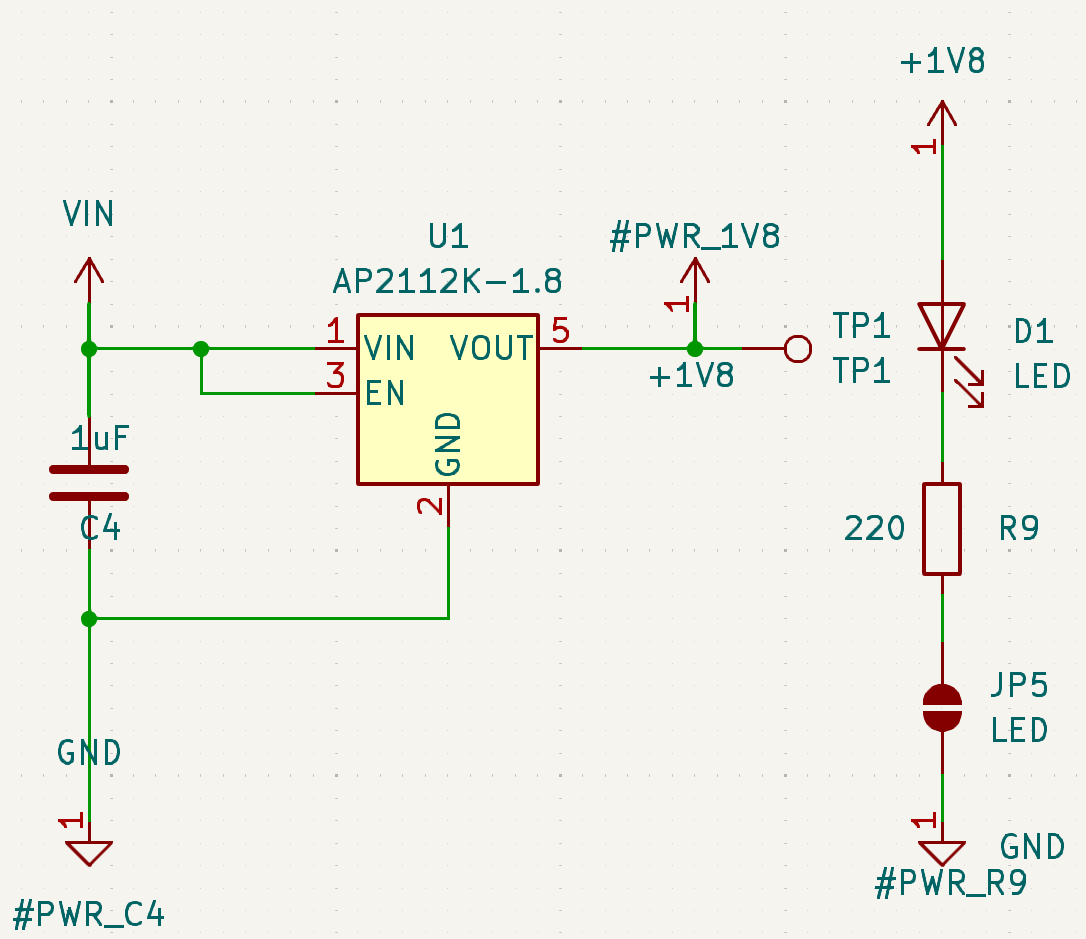}
    \caption{An example schematic designed in \kicad.}
    \label{example_fig}
\end{figure}

Fig.~\ref{example_fig} shows the PCB schematic design of the voltage regulation module and an LED indicator attached, along with three different levels of code representations of the given schematic.

\FloatBarrier
\begin{lstlisting}[style=py, caption={Level 1 Representation}, label={example_code_L1}]
# Auto-generated schematic symbols
import sys
import os

# Get project path and import kicad schematic interface
PROJECT_PATH = os.environ['PROJECT_PATH']
sys.path.append(PROJECT_PATH)
from modules.kicad_sch_interface import *

### Placing center symbol 1 : Regulator_Linear:AP2112K-1.8###

center_x_1, center_y_1 = 120, 105

add_schematic_symbol(symbol_lib="Regulator_Linear", symbol_name="AP2112K-1.8", pos_x=center_x_1, pos_y=center_y_1, reference="U1", value="AP2112K-1.8", rotation=0, mirror="None")

### Placing other symbols in the Schematic with respect to the center symbol 1###

add_schematic_symbol(symbol_lib="power", symbol_name="VAA", pos_x=center_x_1 + (-20), pos_y=center_y_1 + (5), reference="#PWR1", value="VIN", rotation=0, mirror="None")
add_schematic_symbol(symbol_lib="Device", symbol_name="C", pos_x=center_x_1 + (-20), pos_y=center_y_1 + (-5), reference="C1", value="1uF", rotation=0, mirror="None")
add_schematic_symbol(symbol_lib="power", symbol_name="GND", pos_x=center_x_1 + (-20), pos_y=center_y_1 + (-24), reference="#PWR_C1", value="GND", rotation=0, mirror="None")
add_schematic_symbol(symbol_lib="power", symbol_name="+1V8", pos_x=center_x_1 + (13), pos_y=center_y_1 + (5), reference="#PWR_1V1", value="+1V8", rotation=0, mirror="None")
add_schematic_symbol(symbol_lib="Connector", symbol_name="TestPoint", pos_x=center_x_1 + (16), pos_y=center_y_1 + (2), reference="TP1", value="TP1", rotation=270, mirror="x")

### Placing all global labels in the Schematic and connect them to the neighbor pin ###


### Connecting all wires in the Schematic ###


# Connecting #PWR_1V1 pin +1V8 (Pin ID 1 -- Name +1V8) to TP1 pin TP1 (Pin ID 1 -- Name TP1)
connect_pins("#PWR_1V1", "+1V8", "TP1", "TP1")

# Connecting #PWR1 pin VIN (Pin ID 1 -- Name VIN) to C1 pin 1 (Pin ID 1 -- Name None)
connect_pins("#PWR1", "VIN", "C1", "1")

# Connecting U1 pin VOUT (Pin ID 5 -- Name VOUT) to #PWR_1V1 pin +1V8 (Pin ID 1 -- Name +1V8)
connect_pins("U1", "VOUT", "#PWR_1V1", "+1V8")

# Connecting #PWR1 pin VIN (Pin ID 1 -- Name VIN) to U1 pin VIN (Pin ID 1 -- Name VIN)
connect_pins("#PWR1", "VIN", "U1", "VIN")

# Connecting C1 pin 2 (Pin ID 2 -- Name None) to #PWR_C1 pin 1 (Pin ID 1 -- Name None)
connect_pins("C1", "2", "#PWR_C1", "1")

# Connecting U1 pin VIN (Pin ID 1 -- Name VIN) to U1 pin EN (Pin ID 3 -- Name EN)
connect_pins("U1", "VIN", "U1", "EN")

# Connecting C1 pin 2 (Pin ID 2 -- Name None) to U1 pin 2 (Pin ID 2 -- Name None)
connect_pins("C1", "2", "U1", "2")

### Placing center symbol 2 : Device:LED###

center_x_2, center_y_2 = 149, 108

add_schematic_symbol(symbol_lib="Device", symbol_name="LED", pos_x=center_x_2, pos_y=center_y_2, reference="D1", value="LED", rotation=90, mirror="None")

### Placing other symbols in the Schematic with respect to the center symbol 2###

add_schematic_symbol(symbol_lib="power", symbol_name="+1V8", pos_x=center_x_2 + (0), pos_y=center_y_2 + (10), reference="#PWR_1V2", value="+1V8", rotation=0, mirror="None")
add_schematic_symbol(symbol_lib="Device", symbol_name="R", pos_x=center_x_2 + (0), pos_y=center_y_2 + (-11), reference="R1", value="220", rotation=0, mirror="None")
add_schematic_symbol(symbol_lib="Jumper", symbol_name="SolderJumper_2_Open", pos_x=center_x_2 + (0), pos_y=center_y_2 + (-21), reference="JP1", value="LED", rotation=270, mirror="None")
add_schematic_symbol(symbol_lib="power", symbol_name="GND", pos_x=center_x_2 + (0), pos_y=center_y_2 + (-27), reference="#PWR_R1", value="GND", rotation=0, mirror="None")

### Placing all global labels in the Schematic and connect them to the neighbor pin ###


### Connecting all wires in the Schematic ###


# Connecting JP1 pin B (Pin ID 2 -- Name B) to #PWR_R1 pin 1 (Pin ID 1 -- Name None)
connect_pins("JP1", "B", "#PWR_R1", "1")

# Connecting R1 pin 2 (Pin ID 2 -- Name None) to JP1 pin A (Pin ID 1 -- Name A)
connect_pins("R1", "2", "JP1", "A")

# Connecting D1 pin K (Pin ID 1 -- Name K) to R1 pin 1 (Pin ID 1 -- Name None)
connect_pins("D1", "K", "R1", "1")

# Connecting #PWR_1V2 pin +1V8 (Pin ID 1 -- Name +1V8) to D1 pin A (Pin ID 2 -- Name A)
connect_pins("#PWR_1V2", "+1V8", "D1", "A")

write_out_all_wires()


\end{lstlisting}

\begin{lstlisting}[style=py, caption={Level 2 Representation}, label={example_code_L2}]
# Auto-generated schematic symbols
import sys
import os

# Get project path and import kicad schematic interface
PROJECT_PATH = os.environ['PROJECT_PATH']
sys.path.append(PROJECT_PATH)
from modules.kicad_sch_interface import *

### Placing center symbol 1 : Regulator_Linear:AP2112K-1.8###
center_x_1, center_y_1 = 120.650, 104.590
add_schematic_symbol(symbol_lib="Regulator_Linear", symbol_name="AP2112K-1.8", pos_x=center_x_1, pos_y=center_y_1, reference="U1", value="AP2112K-1.8", rotation=0, mirror="None")

### Placing other symbols in the Schematic with respect to the center symbol 1###
add_schematic_symbol(symbol_lib="power", symbol_name="VAA", pos_x=100.33, pos_y=109.67, reference="#PWR1", value="VIN", rotation=0, mirror="None")
add_schematic_symbol(symbol_lib="Device", symbol_name="C", pos_x=100.33, pos_y=99.51, reference="C4", value="1uF", rotation=0, mirror="None")
add_schematic_symbol(symbol_lib="power", symbol_name="GND", pos_x=100.33, pos_y=80.46, reference="#PWR_C4", value="GND", rotation=0, mirror="None")
add_schematic_symbol(symbol_lib="power", symbol_name="+1V8", pos_x=134.62, pos_y=109.67, reference="#PWR_1V8", value="+1V8", rotation=0, mirror="None")
add_schematic_symbol(symbol_lib="Connector", symbol_name="TestPoint", pos_x=137.16, pos_y=107.13, reference="TP1", value="TP1", rotation=270, mirror="x")

### Placing all global labels in the Schematic and connect them to the neighbor pin ###

### Connecting all wires in the Schematic ###
# Connecting #PWR_1V8 pin +1V8 (Pin ID 1 -- Name +1V8) to TP1 pin TP1 (Pin ID 1 -- Name TP1)
connect_pins("#PWR_1V8", "+1V8", "TP1", "TP1")
# Connecting #PWR1 pin VIN (Pin ID 1 -- Name VIN) to C4 pin 1 (Pin ID 1 -- Name None)
connect_pins("#PWR1", "VIN", "C4", "1")
# Connecting U1 pin VOUT (Pin ID 5 -- Name VOUT) to TP1 pin TP1 (Pin ID 1 -- Name TP1)
connect_pins("U1", "VOUT", "TP1", "TP1")
# Connecting U1 pin VIN (Pin ID 1 -- Name VIN) to U1 pin EN (Pin ID 3 -- Name EN)
connect_pins("U1", "VIN", "U1", "EN")
# Connecting C4 pin 2 (Pin ID 2 -- Name None) to #PWR_C4 pin 1 (Pin ID 1 -- Name None)
connect_pins("C4", "2", "#PWR_C4", "1")
# Connecting #PWR1 pin VIN (Pin ID 1 -- Name VIN) to U1 pin VIN (Pin ID 1 -- Name VIN)
connect_pins("#PWR1", "VIN", "U1", "VIN")
# Connecting C4 pin 2 (Pin ID 2 -- Name None) to U1 pin 2 (Pin ID 2 -- Name None)
connect_pins("C4", "2", "U1", "2")

### Placing center symbol 2 : Device:LED###
center_x_2, center_y_2 = 148.590, 108.400
add_schematic_symbol(symbol_lib="Device", symbol_name="LED", pos_x=center_x_2, pos_y=center_y_2, reference="D1", value="LED", rotation=90, mirror="None")

### Placing other symbols in the Schematic with respect to the center symbol 2###
add_schematic_symbol(symbol_lib="power", symbol_name="+1V8", pos_x=148.59, pos_y=118.56, reference="#PWR_1V1", value="+1V8", rotation=0, mirror="None")
add_schematic_symbol(symbol_lib="Device", symbol_name="R", pos_x=148.59, pos_y=96.97, reference="R9", value="220", rotation=0, mirror="None")
add_schematic_symbol(symbol_lib="Jumper", symbol_name="SolderJumper_2_Open", pos_x=148.59, pos_y=86.81, reference="JP5", value="LED", rotation=270, mirror="None")
add_schematic_symbol(symbol_lib="power", symbol_name="GND", pos_x=148.59, pos_y=80.46, reference="#PWR_R9", value="GND", rotation=0, mirror="None")

### Placing all global labels in the Schematic and connect them to the neighbor pin ###

### Connecting all wires in the Schematic ###

# Connecting JP5 pin B (Pin ID 2 -- Name B) to #PWR_R9 pin 1 (Pin ID 1 -- Name None)
connect_pins("JP5", "B", "#PWR_R9", "1")
# Connecting R9 pin 2 (Pin ID 2 -- Name None) to JP5 pin A (Pin ID 1 -- Name A)
connect_pins("R9", "2", "JP5", "A")
# Connecting D1 pin K (Pin ID 1 -- Name K) to R9 pin 1 (Pin ID 1 -- Name None)
connect_pins("D1", "K", "R9", "1")
# Connecting #PWR_1V1 pin +1V8 (Pin ID 1 -- Name +1V8) to D1 pin A (Pin ID 2 -- Name A)
connect_pins("#PWR_1V1", "+1V8", "D1", "A")

write_out_all_wires()
\end{lstlisting}

\clearpage
\begin{lstlisting}[style=py, caption={Level 3 Representation}, label={example_code_L3}]
# Auto-generated schematic symbols
import sys
import os

# Get project path and import kicad schematic interface
PROJECT_PATH = os.environ['PROJECT_PATH']
sys.path.append(PROJECT_PATH)
from modules.kicad_sch_interface import *

### Placing center symbol 1 : Regulator_Linear:AP2112K-1.8###
center_x_1, center_y_1 = 120.650, 104.590
add_schematic_symbol(symbol_lib="Regulator_Linear", symbol_name="AP2112K-1.8", pos_x=center_x_1, pos_y=center_y_1, reference="U1", value="AP2112K-1.8", rotation=0, mirror="None")

### Placing other symbols in the Schematic with respect to the center symbol 1###
add_schematic_symbol(symbol_lib="power", symbol_name="VAA", pos_x=100.33, pos_y=109.67, reference="#PWR1", value="VIN", rotation=0, mirror="None")
add_schematic_symbol(symbol_lib="Device", symbol_name="C", pos_x=100.33, pos_y=99.51, reference="C4", value="1uF", rotation=0, mirror="None")
add_schematic_symbol(symbol_lib="power", symbol_name="GND", pos_x=100.33, pos_y=80.46, reference="#PWR_C4", value="GND", rotation=0, mirror="None")
add_schematic_symbol(symbol_lib="power", symbol_name="+1V8", pos_x=134.62, pos_y=109.67, reference="#PWR_1V8", value="+1V8", rotation=0, mirror="None")
add_schematic_symbol(symbol_lib="Connector", symbol_name="TestPoint", pos_x=137.16, pos_y=107.13, reference="TP1", value="TP1", rotation=270, mirror="x")

### Placing all global labels in the Schematic and connect them to the neighbor pin ###

### Adding all wires in the Schematic ###
add_new_wire([106.68, 107.13], [113.03, 107.13])
add_new_wire([100.33, 91.89], [100.33, 95.7])
add_new_wire([120.65, 91.89], [120.65, 96.97])
add_new_wire([100.33, 80.46], [100.33, 91.89])
add_new_wire([134.62, 109.67], [134.62, 107.13])
add_new_wire([100.33, 103.32], [100.33, 107.13])
add_new_wire([106.68, 104.59], [106.68, 107.13])
add_new_wire([100.33, 107.13], [100.33, 109.67])
add_new_wire([106.68, 104.59], [113.03, 104.59])
add_new_wire([128.27, 107.13], [134.62, 107.13])
add_new_wire([100.33, 91.89], [120.65, 91.89])
add_new_wire([100.33, 107.13], [106.68, 107.13])
add_new_wire([134.62, 107.13], [137.16, 107.13])

### Placing center symbol 2 : Device:R###
center_x_2, center_y_2 = 148.590, 96.970
add_schematic_symbol(symbol_lib="Device", symbol_name="R", pos_x=center_x_2, pos_y=center_y_2, reference="R9", value="220", rotation=0, mirror="None")

### Placing other symbols in the Schematic with respect to the center symbol 2###
add_schematic_symbol(symbol_lib="power", symbol_name="+1V8", pos_x=148.59, pos_y=118.56, reference="#PWR_1V1", value="+1V8", rotation=0, mirror="None")
add_schematic_symbol(symbol_lib="Device", symbol_name="LED", pos_x=148.59, pos_y=108.4, reference="D1", value="LED", rotation=90, mirror="None")
add_schematic_symbol(symbol_lib="Jumper", symbol_name="SolderJumper_2_Open", pos_x=148.59, pos_y=86.81, reference="JP5", value="LED", rotation=270, mirror="None")
add_schematic_symbol(symbol_lib="power", symbol_name="GND", pos_x=148.59, pos_y=80.46, reference="#PWR_R9", value="GND", rotation=0, mirror="None")

### Placing all global labels in the Schematic and connect them to the neighbor pin ###

### Adding all wires in the Schematic ###
add_new_wire([148.59, 104.59], [148.59, 100.78])
add_new_wire([148.59, 83], [148.59, 80.46])
add_new_wire([148.59, 93.16], [148.59, 90.62])
add_new_wire([148.59, 118.56], [148.59, 112.21])

write_out_all_wires()
\end{lstlisting}

\subsection{Example of the request}
\label{example_dataset}

\begin{lstlisting}[style=py, caption={Concise style of request}, label={concise}]
{"messages": [{"role": "user", "content": "I want a 1.8V regulated supply from VIN using an AP2112K LDO, with a test point on the 1.8V rail and a solder-jumper-selectable LED indicator.\n"}]}
\end{lstlisting}

\begin{lstlisting}[style=py, caption={Detailed style of request}, label={detailed}]
{"messages": [{"role": "user", "content": "I am looking to add a small 1.8 V power block to a larger board, mainly to step down a VIN rail and provide a clean +1V8 output. I would also like a simple visual indicator (LED) to show when the rail is active, but with the option to disable it using a solder jumper. For the regulator, please use an AP2112K-1.8 in SOT-23-5. VIN will be the input, and the output should be labeled +1V8. The enable pin should be tied to VIN so the regulator turns on automatically whenever power is present. The NC pin can remain unconnected. On the input side, include a 1 uF capacitor between VIN and GND for decoupling. For the status indicator, I want an LED connected to the +1V8 rail. The idea is that it only lights up when I close a solder jumper. The LED should go from +1V8 through a series resistor (220 Ohm), then into a solder jumper, and finally to GND. If the jumper is open, the LED stays off; if shorted, it lights up. Also, please add a test point on the +1V8 rail so I can easily probe it during debugging. In terms of naming, I would like to keep VIN, +1V8, and GND as the main net labels. The LED path can use intermediate net names generated by the tool.\n"}]}
\end{lstlisting}

\subsection{Dataset Statistics}
\label{dataset:stat}

Here, we define the complexity of any schematic as the sum of the number of symbols and labels.

\begin{figure}[htbp]
    \centering
    \includegraphics[width=0.75\linewidth]{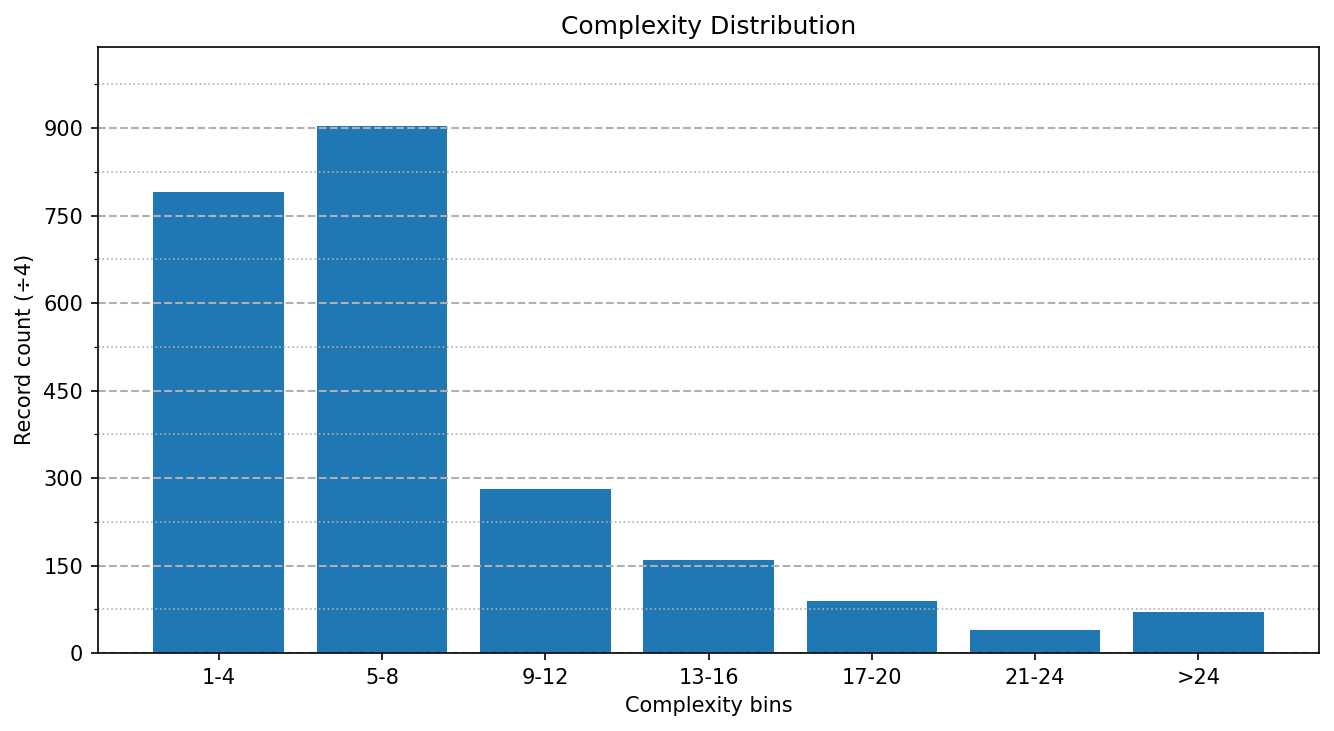}
    \caption{Histogram of the complexity distribution of the dataset.}
    \label{fig:dataset_stat}
\end{figure}

\FloatBarrier
  \subsection{Component Overlap with the GitHub Dataset}
  \label{app:component-overlap}

  We audit component-symbol overlap between the main dataset, including its training and test splits, and the unseen GitHub dataset. After excluding universal resistor, capacitor, and inductor symbols, the main dataset contains 3,302 component instances across 542 unique component identities, while the GitHub dataset contains 985 instances across 394 unique identities.

  As shown in \Cref{tab:component-overlap}, the datasets share 22 unique component identities, corresponding to 5.58\% of the GitHub component vocabulary. Of the 985 GitHub component instances, 56 have identities that also occur in the main dataset, giving an instance-level overlap rate of 5.69\%.

  \begin{table}[htbp]
    \centering
    \small
    \caption{Component-symbol overlap between the main and GitHub datasets after excluding universal resistor, capacitor, and inductor symbols. Overlap percentages use the GitHub counts as denominators.}
    \label{tab:component-overlap}
    \begin{tabular}{lrrr}
      \toprule
      & Main dataset & GitHub dataset & Overlap \\
      \midrule
      Component types & 542 & 394 & 22 (5.58\%) \\
      Component instances & 3,302 & 985 & 56 (5.69\%) \\
      \bottomrule
    \end{tabular}
  \end{table}

  The remaining overlaps mostly consist of commonly used symbols in the \kicad library, such as switches and connectors. These results show that most non-universal component identities and instances in the GitHub dataset are absent from the main dataset, supporting evaluation on a largely different component vocabulary. This component-level audit does not, by itself, rule out overlap in circuit structures or reuse of learned circuit patterns.

\clearpage
\subsection{Training Setup}
\label{app:training}

\begin{table}[htbp]
\centering
\small
\caption{Training configurations for Code-L1 full-parameter supervised fine-tuning. Qwen3.8-27B is fine-tuned on its language-model parameters, with the vision components frozen.}
\label{tab:sft-lora-settings}
\setlength{\tabcolsep}{4pt}
\renewcommand{\arraystretch}{1.12}
\begin{tabular}{@{}>{\raggedright\arraybackslash}m{0.32\linewidth}
                >{\centering\arraybackslash}m{\dimexpr0.34\linewidth-2\tabcolsep\relax}
                >{\centering\arraybackslash}m{\dimexpr0.34\linewidth-2\tabcolsep\relax}@{}}
\toprule
\textbf{Setting} & \textbf{Qwen3.8-27B} & \textbf{GPT-OSS-20B} \\
\midrule
\multicolumn{3}{@{}l}{\textbf{Training setup}} \\
\addlinespace[2pt]
Training method & Full language-model SFT;\newline vision frozen & Full-parameter SFT \\
Assistant-only loss & Yes & Yes \\
Maximum sequence length & 16,384 & 13,312 \\
Sequence packing & Yes & No \\
Epochs & 2 & 2 \\
Microbatch size per device & 1 & 1 \\
Gradient accumulation steps & 8 & 2 \\
Global batch size & 32 packed sequences & 32 examples \\
Gradient checkpointing & Full recomputation & FSDP activation checkpointing \\
Training quantization & None & MXFP4 source,\newline dequantized to BF16 \\
\midrule
\multicolumn{3}{@{}l}{\textbf{Optimization}} \\
\addlinespace[2pt]
Optimizer & Adam (Megatron recipe);\newline CPU offload & Fused AdamW \\
Adam $(\beta_1, \beta_2)$ & $(0.9, 0.98)$ & $(0.9, 0.95)$ \\
Weight decay & 0.1 & 0.1 \\
Learning rate & $2 \times 10^{-5}$ & $2 \times 10^{-5}$ \\
Warmup ratio & 0.05 & 0.03 \\
Learning-rate schedule & Cosine with minimum LR & Cosine with minimum LR \\
Minimum learning rate & $2 \times 10^{-6}$ & $2 \times 10^{-6}$ \\
Minimum learning-rate ratio & 0.1 & 0.1 \\
Gradient clipping & 1.0 & 1.0 \\
\midrule
\multicolumn{3}{@{}l}{\textbf{Compute details}} \\
\addlinespace[2pt]
GPU & NVIDIA A100 80\,GB & NVIDIA H100 80\,GB \\
Number of GPUs & 16 & 16 \\
Parallelism & TP $4$ $\times$ DP $4$ & FSDP full-shard, 16 ranks \\
\bottomrule
\end{tabular}
\par\smallskip
\begin{minipage}{\linewidth}
\footnotesize
TP: tensor parallelism; DP: data parallelism; FSDP: fully sharded data parallelism. BF16 denotes bfloat16. The GPT-OSS-20B scheduler is \texttt{cosine\_with\_min\_lr}.
\end{minipage}
\end{table}

\FloatBarrier
\subsection{Expert Evaluation Rubric}
\label{app:expert-rubric}

Evaluators review the user request, the generated schematic, the reference schematic, and relevant component datasheets when necessary. The reference schematic represents one valid implementation rather than the only acceptable solution; electrically equivalent alternatives that satisfy the request are accepted. Each schematic is assessed using the following three criteria.

\noindent\textbf{Symbol Error.}
Evaluators count missing, incorrect, or duplicated components. When the user request does not specify an exact component, electrically suitable alternatives are accepted and are not counted as errors.

\noindent\textbf{Connection Error.}
Evaluators count required functional pins that are unconnected, connected to an incorrect net, or involved in an unintended short. Electrically equivalent connectivity is accepted regardless of graphical wire routing or net-label naming. Differences in visual appearance alone therefore do not constitute connection errors.

\noindent\textbf{Functional Correctness.}
Evaluators make a binary judgment of whether the complete schematic can reasonably perform the requested functions. Zero Symbol Error and Connection Error counts are necessary but not sufficient for a positive judgment. Evaluators additionally assess end-to-end functionality using their engineering experience and consult relevant datasheets when necessary. This criterion reflects expert assessment of schematic functionality, rather than measured operation of a fabricated board.

Symbol Error and Connection Error are reported as average per-schematic counts over the evaluated schematics. Functional Correctness is reported as the percentage of evaluated schematics receiving a positive judgment.

\subsection{Examples of generation}
\label{sec:example}

\begin{figure}[htbp]
    \centering
    \begin{subfigure}[t]{0.38\linewidth}
        \centering
        \includegraphics[width=\linewidth,height=4.2cm,keepaspectratio]{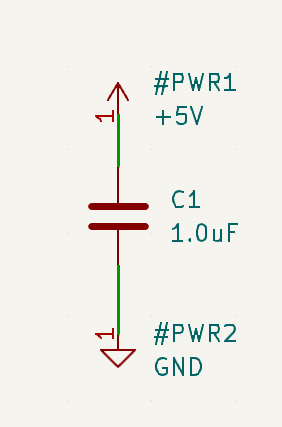}
        \caption{User Request--``I want a 1uF capacitor connected between +5V and GND for power supply decoupling."}
        \label{fig:eg1}
    \end{subfigure}
    \hfill
    \begin{subfigure}[t]{0.58\linewidth}
        \centering
        \includegraphics[width=\linewidth,height=4.2cm,keepaspectratio]{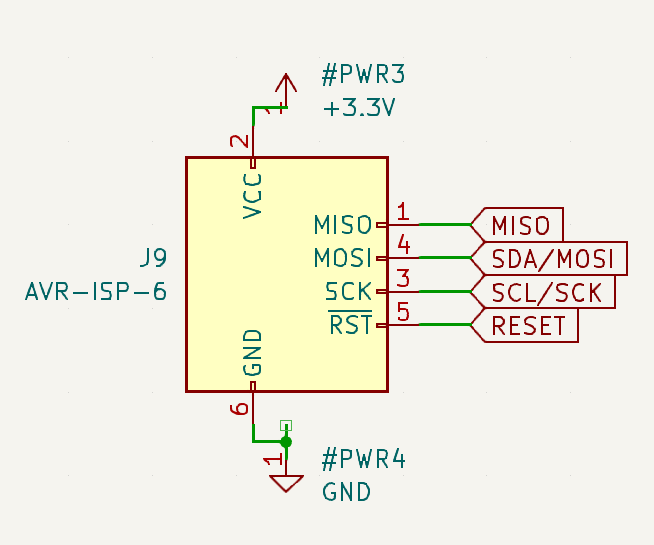}
        \caption{User Request--``I want a 3.3V AVR ISP-6 programming header exposing MISO, MOSI/SDA, SCK/SCL, RESET, VCC, and GND to program a microcontroller."}
        \label{fig:eg2}
    \end{subfigure}
    \caption{Examples of using \name{} to generate schematics based on users' requests.}
    \label{fig:test}
\end{figure}

\begin{figure}[htbp]
    \centering
    \begin{subfigure}[t]{0.38\linewidth}
        \centering
        \includegraphics[width=\linewidth,height=4.5cm,keepaspectratio]{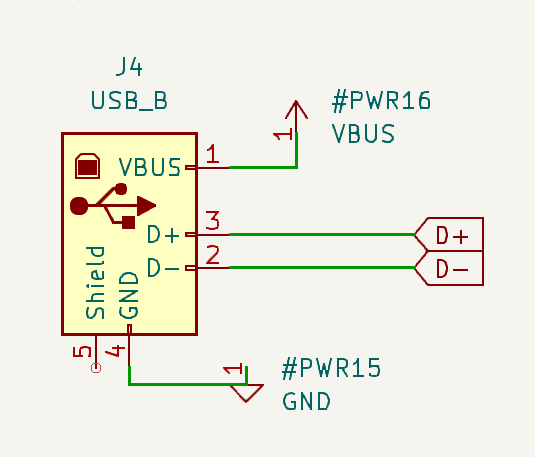}
        \caption{Novel User Request--``I would like to add a USB-B connector interface in the schematic, exporting two labels, namely D+ and D-."}
        \label{fig:eg3}
    \end{subfigure}
    \hfill
    \begin{subfigure}[t]{0.58\linewidth}
        \centering
        \includegraphics[width=\linewidth,height=4.5cm,keepaspectratio]{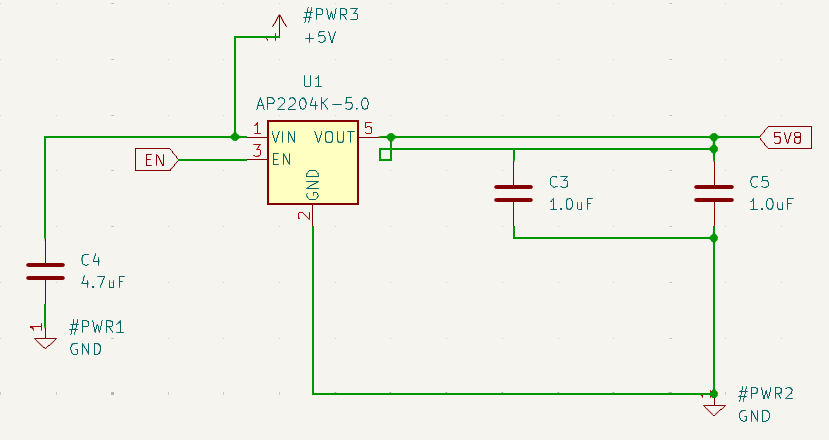}
        \caption{Novel User Request--``I would like a voltage regulator module with a 5V output, using AP2204K."}
        \label{fig:eg4}
    \end{subfigure}
    \caption{Examples of using \name{} to generate a schematic based on users' requests over unseen chips.}
    \label{fig:novel}
\end{figure}

\clearpage
\subsection{Request Styles and Reconstruction Effort}
\label{app:request-style}

\textbf{Request specificity.}
We use concise and detailed requests to cover users with different levels of hardware-design knowledge. Concise requests specify intended functionality, whereas detailed requests additionally constrain components or connections. Requests written by PCB design experts and non-experts are supplied as references when synthesizing the two styles. \Cref{example_dataset} gives examples, and \Cref{tab:request-style} compares their netlist precision.

\begin{table}[htbp]
\centering

\small
\caption{Netlist precision (\%) for concise and detailed requests.}
\label{tab:request-style}
\begin{tabular}{@{}llrr@{}}
\toprule
Model & Representation & Concise & Detailed \\
\midrule
\name{} (Qwen3.8-27B) & Code-L1 & 71.25 & 80.51 \\
\name{} (Qwen3.8-27B) & Code-L2 & 48.42 & 67.92 \\
\name{} (Qwen3.8-27B) & Code-L3 & 34.61 & 33.96 \\
\name{} (GPT-OSS-20B) & Code-L1 & 45.49 & 63.96 \\
GPT-5.5 & Code-L1 & 33.19 & 64.28 \\
GPT-5.5 & Code-L2 & 24.60 & 62.97 \\
GPT-5.5 & Code-L3 & 17.65 & 62.05 \\
\bottomrule
\end{tabular}
\end{table}

\name{} with Qwen3.8-27B and Code-L1 achieves the highest netlist precision for both request styles.

\textbf{Reconstruction effort.}
Approximately 50\% of the initial sketches contain errors, most of which are minor connection issues. On average, fewer than five connections per schematic require adjustment. Human correction and verification take around 10\% of the original design time. More advanced or fine-tuned sketching models can improve reconstruction fidelity.

\subsection{Validation Loss and Generated-Netlist Accuracy}
\label{app:loss-analysis}

The validation loss in \Cref{tab:representation_stat} measures token-prediction loss on the test dataset after training with each representation. Lower validation loss serves as a proxy for better performance and generalization. \Cref{tab:loss-correlations} reports Pearson and Spearman correlations between validation loss and netlist accuracy across samples at three GPT-OSS-20B Code-L1 checkpoints. Both coefficients are statistically significant for Jaccard, precision, and recall at all three checkpoints, with adjusted two-sided $p<0.05$.

\begin{table}[htbp]
\centering

\small
\caption{Correlations between validation loss and netlist accuracy. All $p$-values are adjusted and two-sided.}
\label{tab:loss-correlations}
\setlength{\tabcolsep}{6pt}
\begin{tabular}{@{}llrrrr@{}}
\toprule
Checkpoint & Metric & Pearson $r$ & Adjusted $p$ & Spearman $\rho$ & Adjusted $p$ \\
\midrule
1 & Jaccard & $+0.295$ & $3.16\times10^{-7}$ & $+0.349$ & $1.03\times10^{-13}$ \\
1 & Precision & $+0.266$ & $2.23\times10^{-6}$ & $+0.352$ & $4.33\times10^{-14}$ \\
1 & Recall & $+0.284$ & $6.38\times10^{-7}$ & $+0.348$ & $1.13\times10^{-13}$ \\
\midrule
2 & Jaccard & $+0.234$ & $4.34\times10^{-5}$ & $+0.320$ & $8.96\times10^{-11}$ \\
2 & Precision & $+0.211$ & $2.16\times10^{-4}$ & $+0.308$ & $4.63\times10^{-10}$ \\
2 & Recall & $+0.223$ & $1.01\times10^{-4}$ & $+0.316$ & $1.31\times10^{-10}$ \\
\midrule
3 & Jaccard & $+0.203$ & $8.26\times10^{-4}$ & $+0.271$ & $3.07\times10^{-7}$ \\
3 & Precision & $+0.162$ & $4.31\times10^{-3}$ & $+0.270$ & $3.31\times10^{-7}$ \\
3 & Recall & $+0.181$ & $2.59\times10^{-3}$ & $+0.267$ & $3.71\times10^{-7}$ \\
\bottomrule
\end{tabular}
\end{table}

\subsection{Additional Claude and Gemini Evaluations}
\label{app:additional-models}

We compare \name{} with Claude Sonnet 5, Claude Haiku 4.5, and Gemini 3.8 Flash using Code-L1, Code-L2, Code-L3, and raw \kicad representations. \Cref{tab:additional-models} reports the results. With Code-L1, Qwen-based \name{} reaches 70.7\% netlist Jaccard, compared with 67.95\% for Claude Sonnet 5, 67.04\% for Claude Haiku 4.5, and 63.22\% for Gemini 3.8 Flash.

\subsection{Requirement-Driven Hardware Case Studies}
\label{app:hardware-cases}

\textbf{Case-study setup.}
We conduct two hardware-design case studies involving three users. Each case starts from a natural-language device specification. Users decompose the request into functional blocks, use \name{} to generate block-level schematics, and merge them into a complete design. The fabricated boards are then tested for correctness. We also ask three experts to perform comparable complete-board design tasks and record the time consumed.

\begin{table}[htbp]
\centering

\small
\caption{Hardware design case studies and design times.}
\label{tab:hardware-cases}
\setlength{\tabcolsep}{4pt}
\begin{tabular}{@{}>{\raggedright\arraybackslash}p{0.29\linewidth}
                  >{\raggedright\arraybackslash}p{\dimexpr0.41\linewidth-2\tabcolsep\relax}
                  >{\centering\arraybackslash}p{\dimexpr0.15\linewidth-2\tabcolsep\relax}
                  >{\centering\arraybackslash}p{\dimexpr0.15\linewidth-2\tabcolsep\relax}@{}}
\toprule
Device requirement & Fabricated-hardware check & SchGen time (min) & Mean manual time (min) \\
\midrule
Multi-sensor module: pressure, light, audio, inertial, magnetic, and environmental sensing; I2C and a power-indicator LED &
Gyroscope responses to controlled motion along all three axes. &
15 & 82 \\
\addlinespace
CO$_2$ monitor: sensing, microcontroller processing, USB communication, and microSD storage &
Response to human breath and continuous indoor CO$_2$ recording over 12 hours. &
24 & 128 \\
\bottomrule
\end{tabular}
\end{table}

These cases show how automating repetitive design operations allows human effort to focus on requirements, inspection, correction, and functional verification.

\textbf{Larger schematics.}
Large PCB schematics can be partitioned into functional blocks connected by labels. \name{} can generate the blocks separately and merge them into a complete schematic. Training directly on larger schematics is a further direction.

\subsection{Library and EDA-Tool Portability}
\label{app:portability}

\name{} is a general method for PCB schematic generation based on tool-independent operations: anchor-relative component placement and named-pin connections. We choose KiCad for its free license, open-source implementation, accessible libraries, and file format. The same representation can be transferred to Cadence or Altium by implementing a backend through the target tool's supported scripting interface or SDK.

At inference time, the available library information, including symbol identifiers and pin definitions, is supplied as context. This context can be replaced with metadata from another library, while the backend resolves the supplied identifiers and pin mappings. With compatible library metadata and a backend implementing the same operations, a new library can be used without retraining the model.

\subsection{API Checks, Expert Agreement, and Failure Cases}
\label{app:checks-failures}

\textbf{Execution, electrical checks, and function.}
Code-L1 and Code-L3 are evaluated on the same test requests, with one generation attempt per request and the requirement of zero KiCad ERC errors. Code-L1 checks named pins before constructing a connection, whereas Code-L3 can place wire segments directly at arbitrary coordinates. In \Cref{tab:code-eval}, Code-L1 achieves 70.7\% netlist Jaccard versus 19.6\% for Code-L3, despite its stricter API checks.

\textbf{Expert agreement.}
Six experts assess generated schematics using the user request, reference schematic, and relevant component datasheets, following \Cref{app:expert-rubric}. Electrically equivalent implementations are accepted, as are suitable component substitutes when no exact part is requested. Symbol and connection errors are reported as mean counts per schematic across experts. Functional correctness requires zero symbol and connection errors and a separate assessment of end-to-end functionality, and is reported as the percentage of schematics receiving positive judgments from all experts. Disagreements are reconciled separately. \Cref{tab:expert-agreement} shows the raw disagreement distribution.

\begin{table}[htbp]
\centering

\small
\caption{Expert disagreement distribution over 500 schematics.}
\label{tab:expert-agreement}
\begin{tabular}{@{}rr@{}}
\toprule
Number of disagreements & Number of schematics \\
\midrule
0 & 478 \\
1 & 11 \\
2 & 8 \\
3 & 3 \\
\midrule
Total & 500 \\
\bottomrule
\end{tabular}
\end{table}

\textbf{Failure-case study.}
\Cref{tab:failure-cases} summarizes four failure cases from retained Code-L1 evaluation records for the two \name{} backbones.

\begin{table}[htbp]
\centering

\small
\caption{Failure cases in \name{} Code-L1 generation.}
\label{tab:failure-cases}
\setlength{\tabcolsep}{5pt}
\begin{tabular}{@{}>{\raggedright\arraybackslash}p{0.30\linewidth}
                  >{\raggedright\arraybackslash}p{\dimexpr0.70\linewidth-2\tabcolsep\relax}@{}}
\toprule
Failure and example & Observed evidence \\
\midrule
Incomplete program.\newline Qwen3.8-27B, ZED-F9P GPS &
Generated code ends inside an \texttt{add\_schematic\_symbol} call; a \texttt{SyntaxError} prevents netlist export. \\
\addlinespace
Invalid pin reference.\newline Qwen3.8-27B, indoor air-quality sensor &
The code requests \texttt{LX} on the AP3012 symbol, whose library pin is \texttt{SW}; pin lookup fails before export. \\
\addlinespace
Ambiguous pin selection.\newline Qwen3.8-27B, mosaic-T UART protection &
All four UART signals use \texttt{D1.K}; pins 1, 3, and 4 remain unconnected. Only 2 of 5 reference nets match: precision is 100\%, but recall is 40\%. \\
\addlinespace
Connectivity mismatch after export.\newline GPT-OSS-20B, XBee development board &
Execution/export succeeds, but the single evaluated net matches neither of the two reference nets; precision and recall are both 0\%. \\
\bottomrule
\end{tabular}
\end{table}

\FloatBarrier

\end{document}